\documentclass{openmoss}

\graphicspath{{assets/}{./}}
\usepackage{helvet}

\usepackage{amsmath}
\usepackage{natbib}
\usepackage{graphicx}
\usepackage{subcaption}

\usepackage[utf8]{inputenc}
\usepackage[T1]{fontenc}
\usepackage{hyperref}
\usepackage{url}
\usepackage{booktabs}

\usepackage{amsfonts}
\usepackage{nicefrac}
\usepackage{microtype}
\usepackage{wrapfig}
\usepackage{amssymb}

\usepackage{titletoc}
\usepackage{minitoc}

\usepackage{array}
\usepackage{etoolbox}

\definecolor{lightblue}{RGB}{200, 230, 255}
\definecolor{headerblue}{RGB}{150, 200, 255}

\usepackage{pgfplots}
\pgfplotsset{compat=1.18}
\usepackage{xcolor}
\usepackage{float}
\usepackage{comment}
\usepackage{multirow}
\usepackage{makecell}
\usepackage{siunitx}
\usepackage{tikz}
\usepackage{pgf-pie}
\usepackage[export]{adjustbox}

\usepackage{ragged2e}
\usepackage{tabularx}
\usepackage{caption}
\usepackage{enumitem}
\usepackage{pifont}
\usepackage[hang,flushmargin]{footmisc}

\usepackage{tcolorbox}
\tcbuselibrary{breakable}
\tcbuselibrary{skins}

\usepackage{listings}

\usepackage{threeparttable}
\usepackage{setspace}

\usepackage[scheme=plain,fontset=none]{ctex}
\usepackage{tabularx}

\newcolumntype{Y}{>{\raggedright\arraybackslash}X}
\newcolumntype{P}[1]{>{\raggedright\arraybackslash}p{#1}}

\newcolumntype{L}[1]{>{\RaggedRight\arraybackslash}p{#1}}
\newcolumntype{C}[1]{>{\centering\arraybackslash}p{#1}}
\ifdefined\DeclareUnicodeCharacter
\DeclareUnicodeCharacter{00E9}{\'{e}}
\DeclareUnicodeCharacter{00F6}{\"{o}}
\DeclareUnicodeCharacter{0117}{\.{e}}
\DeclareUnicodeCharacter{017D}{\v{Z}}
\DeclareUnicodeCharacter{2013}{--}
\DeclareUnicodeCharacter{2014}{---}
\DeclareUnicodeCharacter{2018}{\textquoteleft}
\DeclareUnicodeCharacter{2019}{\textquoteright}
\fi

\usepackage[table]{xcolor}

\definecolor{oursgray}{gray}{0.95}

\definecolor{MossCyan}{HTML}{82D9FF}
\definecolor{MossBlue}{HTML}{82B1FF}

\definecolor{lightblue}{RGB}{200, 230, 255}
\definecolor{headerblue}{RGB}{150, 200, 255}
\definecolor{oursgray}{gray}{0.95}
\definecolor{MossCyan}{HTML}{82D9FF}
\definecolor{MossBlue}{HTML}{82B1FF}
\definecolor{tickG}{HTML}{00C853}
\definecolor{crossR}{HTML}{FF1744}
\definecolor{SCIONBar}{HTML}{0B3D91}
\definecolor{DRClawBar}{HTML}{B65A4A}
\definecolor{AutoResearchBar}{HTML}{2F7D6D}
\definecolor{ScienceClawBar}{HTML}{4A7FA7}
\definecolor{AIResearcherBar}{HTML}{7A6A58}
\definecolor{AIScientistBar}{HTML}{A38B32}
\definecolor{EvoScientistBar}{HTML}{C97941}
\definecolor{InternAgentBar}{HTML}{5A5F66}
\definecolor{ARISBar}{HTML}{7D5A8C}

\definecolor{tickG}{HTML}{00C853}
\definecolor{crossR}{HTML}{FF1744}

\newcommand{\faGithub}{\raisebox{-0.2ex}{\includegraphics[height=2.0ex]{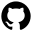}}}

\newtcolorbox{promptbox}[2][]{
    colback=white,
    coltext=black,
    arc=3mm,
    boxrule=0.5pt,
    colframe=black!60!white,
    title={#2},
    colbacktitle=black,
    coltitle=white,
    fonttitle=\bfseries,
    top=8pt,
    bottom=8pt,
    left=10pt,
    right=10pt,
    breakable,
    before upper={%
        \linespread{1}\selectfont
        \setlength{\parskip}{1ex plus 0.2ex minus 0.2ex}%
        \setlength{\parindent}{0pt}%
    },
    #1
}
\renewcommand{\headerlogo}{%
  \includegraphics[height=9mm]{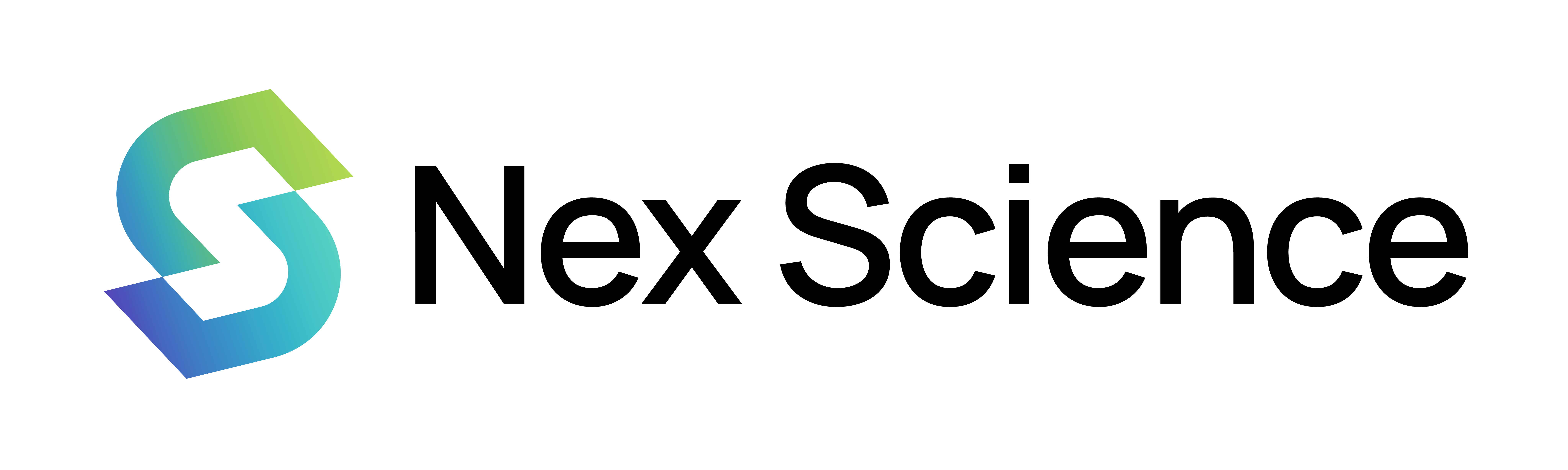}%
}
\title{Rethinking Scientific Discovery in the Agentic Era}

\author{
Yining Zheng$^{1,2,*,\dagger}$,
Yuxin Wang$^{1,2,*,\dagger}$,
Jiahao Lu$^{1,2}$,
Shicheng Fang$^{1,2}$,
Weiyi Wang$^{1}$,
Yongzhuo Yang$^{1,2}$,
Bowen Li$^{1,2}$,
Haochen Ma$^{1,2}$,
Chen Hu$^{1,2}$,
Bowen Chen$^{1,2}$,
Yang Wang$^{1,2}$,
Huanhui Chen$^{1}$,
Yitong Chen$^{1}$,
Jiajun Chen$^{1,2}$,
Zhiyuan Li$^{1,2}$,
Yanlin Li$^{1,2}$,
Zhuo Yang$^{1}$,
Qifeng Wu$^{1,2}$,
Jiaying He$^{1,2}$,
Zhijie Jinluo$^{1,2}$,
Xiaohu Xu$^{1,3}$,
Yi Feng$^{1,2}$,
Juncheng Qian$^{1,2}$,
Yizhou Chen$^{2}$,
Yang Cheng$^{2}$,
Tong Zhu$^{1,3}$,
Tianlei Ying$^{1,2}$,
Hongyu Yu$^{2}$,
Hongjun Xiang$^{2}$,
Xipeng Qiu$^{1,2,\dagger}$\\[2mm]
{\normalfont \normalsize $^{1}$Shanghai Innovation Institute},
{\normalfont \normalsize $^{2}$Fudan University},
{\normalfont \normalsize $^{3}$East China Normal University}\\[1mm]
{\normalfont \normalsize $^{*}$These authors contributed equally.}\\
{\normalfont \normalsize $^{\dagger}$Corresponding authors: 
zhengyining@sii.edu.cn, 
wangyuxin@sii.edu.cn, 
qiuxipeng@sii.edu.cn}
}

\abstract{

Artificial intelligence has achieved remarkable progress in scientific discovery, yet most AI4Science systems remain fragmented tools that still rely on human scientists to coordinate problem formulation, literature grounding, model invocation, simulation, validation, and knowledge reuse. This paper presents \textbf{SCION (Scientific Collaborative Innovation with Agentic Organizational Nexus)}, an agentic scientific operating system that serves as an \textbf{organizational nexus}: through a Science Agent acting as a \textbf{Meta-Harness}, it connects scientific tasks, tools, agents, artifacts, and memory across a collaborative research organization. Rather than functioning as a single model, chatbot, or workflow script, SCION transforms scientific research itself into an executable, auditable, and reusable operational process. The system centers on the \textbf{Research Execution Plan (REP)}, a structured representation that compiles high-level scientific intent into staged objectives, dependencies, verification checkpoints, tool requirements, expected artifacts, and fallback conditions. On this basis, SCION integrates a hierarchical multi-agent runtime, profile-driven agent specialization, selective context construction, governed delegation, and layered epistemic memory to support long-horizon scientific execution. We further formulate scientific discovery under SCION as \textbf{Target-conditioned Inverse Search}, and extend this formulation to hidden-target settings through batch active search under finite experimental budgets. These formulations show how the Science Agent's Meta-Harness converts scientific goals, constraints, and feedback into adaptive search procedures over candidates, tools, and experimental trajectories. Representative applications are discussed in materials analysis, multi-property molecule design, and target-specific protein or antibody screening. Experiments on scientific reading, idea generation, molecule generation, and antibody screening show that SCION improves over existing autonomous research-agent baselines, particularly in tasks requiring decomposition, verification, iterative refinement, and memory reuse. Overall, SCION shifts AI from isolated predictive tools toward a coordinated operational layer---an organizational nexus---for scientific work, enabling more traceable, recoverable, and reusable modes of AI-assisted scientific innovation. 
}

\checkdata[\raisebox{-0.1ex}{\faGithub}\hspace{0.4em}GitHub]{\url{https://github.com/OpenMOSS/ScienceEvolve}}

\begin{document}
\maketitle
\begingroup
\renewcommand{\thefootnote}{\fnsymbol{footnote}}
\setcounter{footnote}{0}
\footnotetext[1]{These authors contributed equally.}
\footnotetext[2]{Corresponding authors.}
\endgroup

\section{Introduction}
Over the past decade, artificial intelligence has become an important driver of scientific discovery. In the AI for Science (AI4Science) paradigm, data-driven and deep-learning methods have been used for structure prediction, property estimation, candidate generation, literature analysis, and experimental planning \cite{wang2023scientific,jumper2021highly,watson2023novo,skarlinski2024languageagentsachievesuperhuman}. Landmark systems such as AlphaFold for protein structure prediction \cite{jumper2021highly} and GNoME for inorganic materials discovery \cite{merchant2023scaling} show that advanced AI models can solve scientific subtasks that were previously difficult to scale with manual or purely physics-based approaches.

However, the success of these models has not yet produced a corresponding transformation of the full research operating model \cite{shin2025revolutionscientificworkflowsagentic}. In most laboratories and research organizations, AI tools are still deployed as isolated capability nodes within a workflow whose coordination remains human-centered \cite{lu2026towards,scheurer2025role}. A scientist must formulate a problem, identify relevant literature, select tools, prepare inputs, launch simulations or model calls, inspect failures, reconcile inconsistent outputs, and decide how each intermediate result should affect the next step. As the number of tools, models, datasets, and experimental options grows, this manual coordination layer becomes a bottleneck \cite{li2025review,kim2025towards}. The limiting factor is no longer only model accuracy, but the ability of the research system to represent intent, coordinate action, preserve memory, and recover from failure across long multi-stage workflows.

Scientific productivity depends not only on stronger models, but also on a new systems layer for scientific work. A useful AI4Science infrastructure should not merely answer questions or predict properties. It should be able to translate high-level research objectives into executable task structures, expose the appropriate tools and context to specialized agents, monitor and validate intermediate outputs, retain both successful and failed trajectories, and make the resulting process auditable by human scientists. In other words, the research workflow itself should become a first-class computational object.

The \textbf{SCION (Scientific Collaborative Innovation with Agentic Organizational Nexus)} is designed around this objective. It is positioned as an agentic organizational nexus for scientific workflows, connecting high-level research intent with the tools, agents, artifacts, and memory of a research group, rather than a single model, chatbot, or visualization dashboard. Its central abstraction is the Research Execution Plan (REP), a structured and machine-operable research plan compiled from high-level scientific intent. A REP is an executable plan that makes the intended research procedure explicit: it specifies research objectives, constraints, planned stages, stage dependencies, verification checkpoints, required tools, expected artifacts, and fallback conditions. It also records how intermediate evidence should affect subsequent actions, so that the system can branch, retry, terminate, or escalate work without losing provenance. On top of this representation, the system coordinates a hierarchy of agents for reading, scientific reasoning, execution, collaboration, and report generation. Beneath it, a layered memory subsystem preserves project history, provenance, intermediate artifacts, failure records, and reusable domain knowledge.

 This paper makes four main contributions. First, it identifies the structural bottleneck created when humans act as the central dispatcher across fragmented AI4Science tools. Second, it describes a three-layer architecture for intent representation, multi-agent execution, and epistemic memory. Third, it formulates scientific discovery under this architecture as target-conditioned inverse search and, under hidden target property condition, as batch active search. Fourth, it discusses representative application patterns in materials analysis, multi-property molecule design, and target-specific antibody screening.

\section{Related Work}

\subsection{Common AI4Science}
Common AI4Science has produced highly capable models for specific scientific tasks. Representative examples include AlphaFold for protein structure prediction \cite{jumper2021highly}, GNoME for inorganic materials discovery \cite{merchant2023scaling}, and data-driven models for mappings such as structure-to-property \cite{xie2018crystal}, composition-to-property \cite{dunn2020benchmarking}, property estimation, candidate generation, literature analysis, and experimental planning \cite{wang2023scientific,watson2023novo,skarlinski2024languageagentsachievesuperhuman}. Together, this line of work shows that learning-based methods can outperform manual or purely physics-based approaches on important local problems, making them indispensable components of modern AI4Science workflows.

However, these advances do not by themselves transform the full scientific research process. Most common AI4Science models are still optimized for isolated input-output tasks, while scientific discovery is a cross-stage workflow involving problem formulation, literature grounding, hypothesis decomposition, tool selection, simulation or experiment design, anomaly handling, validation, and knowledge reuse \cite{shin2025revolutionscientificworkflowsagentic,wei2025ai}. As a result, their outputs often remain passive references for human decision-making rather than active components of a coordinated research runtime. Human scientists must still act as the central dispatcher across fragmented tools, databases, software environments, and analytical platforms \cite{lu2026towards,scheurer2025role}. The resulting bottleneck is not only model accuracy, but the absence of a system-level infrastructure for representing research intent, coordinating multi-stage execution, preserving memory and provenance, and recovering from failure across long-horizon workflows \cite{li2025review,kim2025towards,hysmith2024future}.

\subsection{Agents for Scientific Research}
The rise of Large Language Models (LLMs) has motivated a second line of work that moves beyond isolated prediction models toward agentic implementations for scientific research. Tool-using scientific agents such as ChemCrow \cite{bran2023chemcrow} and Coscientist \cite{boiko2023autonomous} demonstrate that LLMs can plan, invoke domain tools, and maintain intermediate reasoning traces. Pipeline-based autonomous research agents go further by automating larger portions of the research loop, including AI-Researcher \cite{tang2025airesearcherautonomousscientificinnovation}, AI-Scientist-v1 \cite{lu2026towards}, AI-Scientist-v2 \cite{yamada2025aiscientistv2workshoplevelautomated}, InternAgent-1.0 \cite{internagentteam2025internagentagentscientist}, and AutoResearchClaw \cite{liu2026autoresearchclaw}. They typically arrange literature review, idea generation, experiment design, coding, evaluation, and report writing into staged workflows, extending automation from local scientific tasks to broader research procedures.

Agentic research systems introduce more flexible decomposition, tool invocation, coding assistance, retrieval, critique, and iterative refinement. Representative examples include EvoScientist \cite{lyu2026evoscientistmultiagentevolvingai}, DR-Claw \cite{song2026drclaw}, and ScienceClaw \cite{wang2026autonomousagentscoordinatingdistributed}, spanning self-evolving multi-agent frameworks, research workspaces, and distributed artifact exchange. ARIS \cite{yang2026arisautonomousresearchadversarial} represents a related collaborative system built around coding agents such as Claude Code and Codex. These systems highlight the value of multi-agent collaboration and adaptive execution in scientific tasks. Still, most existing agentic implementations remain organized as pipelines, scripts, or task-level collaboration frameworks. Unlike general LLM-agent operating systems \cite{mei2024aios}, they generally lack unified operating-system-level orchestration, explicit lifecycle management, persistent long-term memory \cite{zhang2025survey}, robust recovery mechanisms, and governed provenance tracking. As a result, prolonged tasks can still suffer from hallucinations \cite{liu2026agenthallu}, silent failure propagation, and logical dead loops \cite{mohammadi2025evaluation}. SCION is positioned to address this gap by treating the scientific workflow itself as the primary computational object, with explicit research plans, governed multi-agent execution, layered memory, verification checkpoints, and reusable project-level state.

\section{Rethinking AI and Human Positions in an Agentic Era}

\begin{figure}[htbp]
    \centering
     \includegraphics[width=0.85\linewidth]{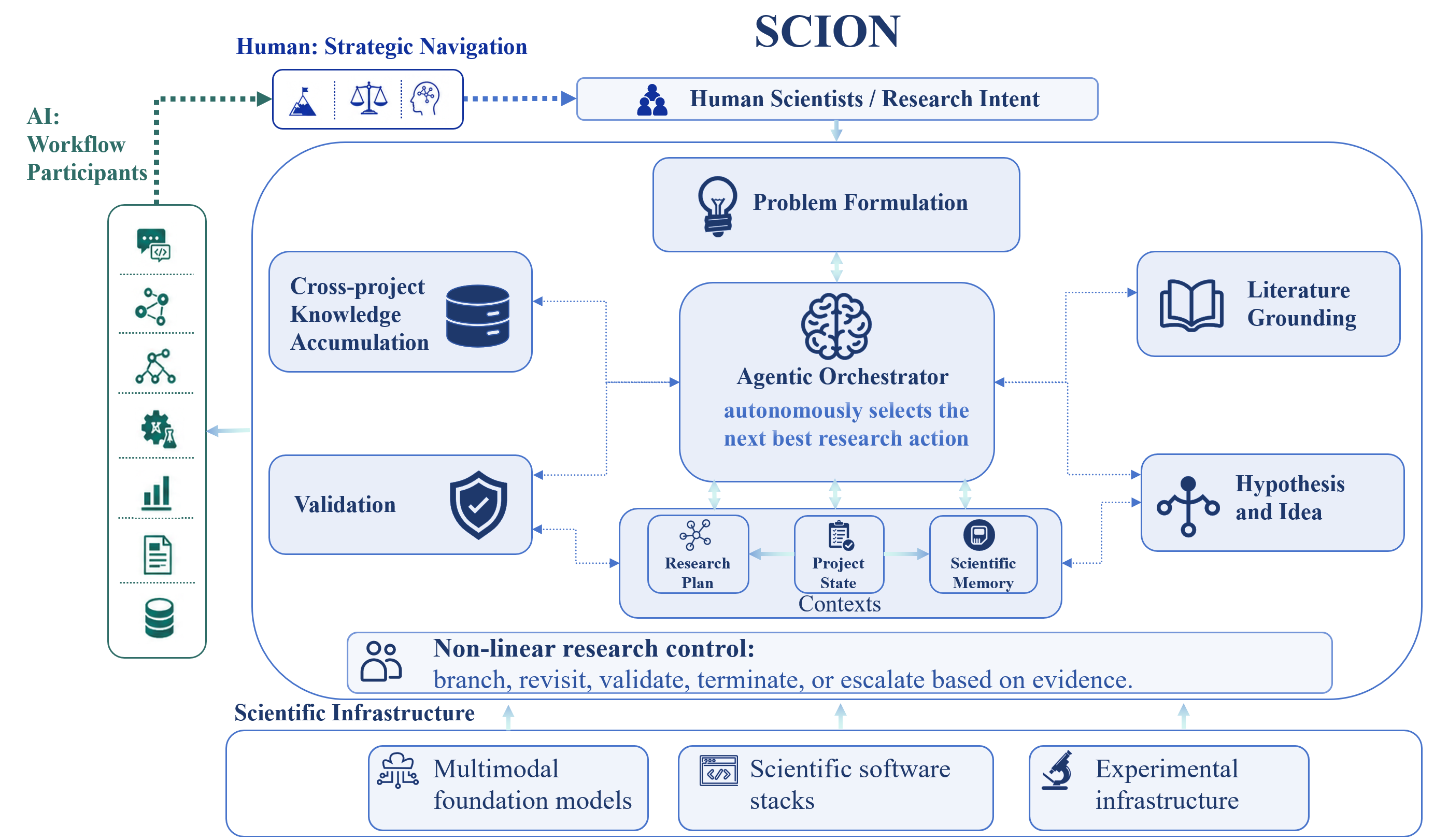}
    \caption{SCION in the organizational research process. The diagram places SCION as the nexus between high-level scientific intent and the heterogeneous tools, agents, artifacts, and memory used during research execution. Its role is to coordinate organizational workflows for laboratories and research groups, rather than merely provide personal assistance or access to an isolated model.}
    \label{fig:conceptandposition}
\end{figure}

SCION is not designed as a visualization dashboard or a monolithic scientific application, but as an \textbf{agentic nexus} that orchestrates the entire scientific research workflow, connecting scientific intent with the tools, agents, artifacts, and memory used to act on it. Unlike many AI-enabled research assistants currently available, which primarily function as personal productivity tools for individual scientists, SCION is designed as an \textbf{organizational solution} for collective research. As summarized in Figure~\ref{fig:conceptandposition}, its target is not only the single-user research loop, but the coordinated workflow of laboratories, research groups, and institutional research programs.

The core objective of SCION is to support structured research agency in AI-assisted workflows. It integrates research goals, autonomous actions, and empirical validations into a unified systems framework, making the research process itself an engineering object that can be compiled, scheduled, monitored, audited, and optimized. In this sense, it is intended to cover the full research lifecycle, including problem formulation, literature grounding, hypothesis decomposition, computational or experimental execution, result validation, and cross-project knowledge accumulation. Its systemic positioning is analogous to an operating system in computer science: it bridges heterogeneous multimodal foundation models, scientific software stacks, and experimental infrastructure at the bottom, while interfacing with the abstract research intents of human scientists at the top.

\subsection{A Meta-Harness Way to Scientific Discovery}

SCION is not intended to remove human scientists from the research process. Rather, it reorganizes the operational division of labor in scientific workflows. In the conventional paradigm, human scientists not only define scientific questions but also function as the runtime layer of the research system: they translate intent into tool calls, move information across incompatible software environments, monitor intermediate failures, and decide when to retry, branch, or terminate an exploration path. As a result, scientific throughput is often constrained not only by model capability, compute, or instrumentation, but also by the human effort required to coordinate them.

Following the broader notion of a \textit{Meta-Harness}, we use the term here to characterize an orchestration layer above individual scientific tools, models, agents, artifacts, and memory systems. A conventional harness connects a model, tool, or procedure to a task-specific workflow. A meta-harness operates one level higher: it coordinates multiple task-level harnesses and execution components into a unified research process. Under this view, the Science Agent inside SCION functions as the meta-harness for scientific discovery. SCION provides the operating-system-level substrate, including plan representation, context construction, memory, delegation, tool boundaries, and governance policies, while the Science Agent composes these resources into executable discovery procedures.

\subsection{AI: From Isolated Tools to Workflow Participants}
Within SCION, AI is repositioned from a passive tool for isolated commands to a stage-aware operational collaborator for scientific work. Guided by the Research Execution Plan, the current project state, and accumulated memory, agents do not mechanically execute a fixed end-to-end pipeline; instead, they judge whether the active work concerns problem formulation, literature grounding, hypothesis decomposition, execution, validation, or cross-project learning, and then activate the appropriate tools, agents, context, and verification mechanisms. In this position, AI can translate high-level intent into executable representations, connect evidence sources to downstream tasks, expand scientific questions into testable branches, coordinate computational or experimental actions, synthesize intermediate evidence, and produce milestone outputs. It also functions as a team-level trajectory manager: assumptions, decision points, artifacts, failed attempts, successful results, and reasoning traces are summarized into reusable scientific memory, enabling research groups to compare branches, decide whether to continue or terminate a path, and transfer structured experience across projects.

\subsection{Humans: From Process Execution to Strategic Navigation}
Human scientists are correspondingly relieved from acting as the glue code of the research system. Their role is elevated toward:
\begin{enumerate}
    \item \textbf{Posing Paradigm-level Questions:} Identifying the most valuable scientific objectives, defining search spaces, and setting the overall direction of investigation.
    \item \textbf{Value Alignment and Ethical Boundary Setting:} Establishing risk thresholds for exploration, including biosafety, toxicity, and scientific compliance constraints.
    \item \textbf{High-dimensional Scientific Judgment:} Intervening when parallel exploration paths require mechanistic interpretation, model trust calibration, or a strategic change of direction.
\end{enumerate}

\section{Multi-Agent Architecture Design}

\begin{figure}[htbp]
    \centering
    \includegraphics[width=0.9\linewidth]{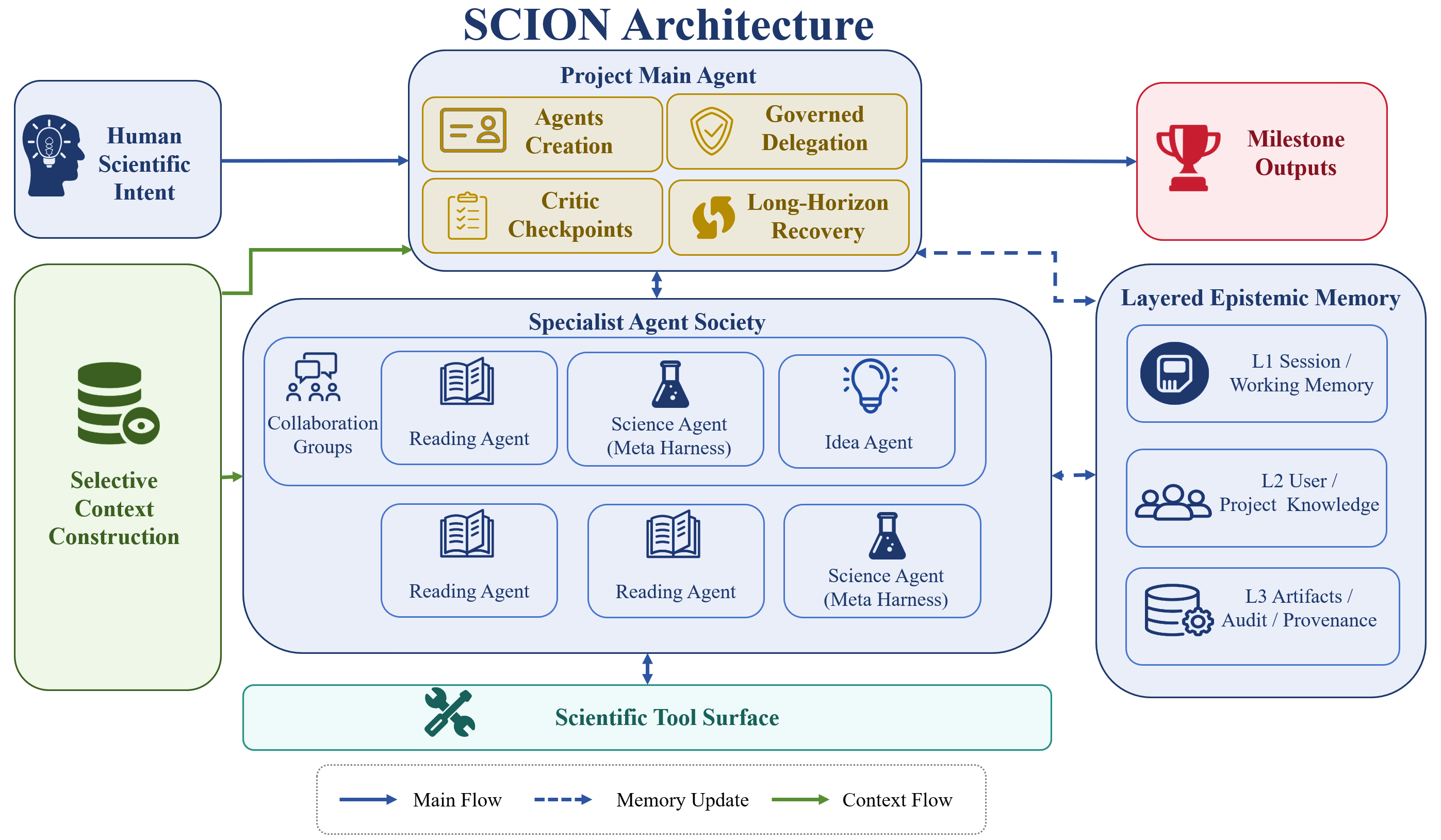}
    \caption{Schematic diagram of the SCION core architecture. The architecture combines a hierarchical multi-agent runtime, selective context construction, layered epistemic memory, and governed long-horizon execution. The diagram highlights how planning, delegation, verification, and memory are coupled so that scientific work can remain auditable and recoverable across extended workflows.}
    \label{fig:architecture}
\end{figure}

The operational core of SCION is best understood through four tightly coupled mechanisms: hierarchical multi-agent organization, profile-driven agent realization, context-governed execution, and layered epistemic memory with recoverable delegation, as illustrated in Figure~\ref{fig:architecture}. Together these mechanisms transform a general-purpose language model into a persistent scientific runtime that can decompose work, invoke heterogeneous tools, preserve continuity, and remain auditable across extended research cycles. Importantly, the system is not implemented as a collection of unrelated assistants. Instead, it uses one common agent kernel and differentiates behavior by agent profiles, memory permissions, tool surfaces, and deployment scope. This makes the architecture simultaneously modular at the level of scientific labor and uniform at the level of runtime realization.

\subsection{Hierarchical Multi-Agent Organization}
Scientific work involves heterogeneous cognitive modes that should not compete within a single reasoning stream. SCION therefore organizes agents as a hierarchy rather than as a flat pool or a monolithic assistant. A central motivation for this organization is context hygiene in open-ended projects. Complex scientific and engineering projects rarely begin with complete specifications; they usually start from rough ideas and become clearer only through discussion, trial, failure, review, and revision. If all of this material remains inside one long interaction, the main session is quickly contaminated by requirement negotiation, implementation details, abandoned attempts, failed tool traces, lengthy explanations, and review records. The session then loses its identity: neither the scientist nor the agent can reliably distinguish global project logic from task-local noise.

SCION therefore treats \texttt{project-main} as a clean project-level master session. It preserves only the global objective, key decisions, branch structure, approved summaries, and process-level documentation needed to keep the project intelligible. Concrete scientific work is delegated to specialist subagents, whose local sessions may contain noisy exploration, clarification questions, failed attempts, and detailed tool traces. Their outputs are reintegrated into \texttt{project-main} only after normalization and summarization. This design avoids two common failure modes in long projects: asking every question inside an already polluted main context, or opening a new session that forces the user to redescribe the entire project from scratch. The \texttt{reading} agent is responsible for literature retrieval, document reading, evidence synthesis, and citation-grounded summaries. The \texttt{idea} agent transforms field background and grounded literature evidence into candidate research ideas, novelty judgments, proposal-level rationales, and prioritized ideation branches. The \texttt{sci} agent handles hypothesis refinement, mechanism-level reasoning, experiment design, and domain-constrained scientific synthesis. The \texttt{exec} agent operates computational environments, scripts, external tools, simulations, and artifact-producing workflows. The \texttt{lark} agent connects the project to collaboration platforms, group communication, and external coordination. These specialists are deliberately task-local: they absorb the detailed operational burden of a branch, including speculative ideation and failed exploration, and return compressed evidence, decisions, and artifacts to the project-level coordinator. Furthermore, collaboration groups can be established to address milestone-related challenges by facilitating focused discussion, cross-functional collaboration, and timely problem-solving. At each moment, the active agent set is denoted by
\begin{equation}
\mathcal{A}_t = \{a_c, a_1, a_2, \dots, a_k\},
\end{equation}
where $a_c$ is the coordinator agent (project-main agent) and $a_1, \dots, a_k$ are specialist agents responsible for literature grounding, idea generation and evaluation, scientific reasoning, tool execution, and collaboration-platform interaction.
The coordinator maintains project-level continuity, decides whether to respond directly or delegate, and integrates downstream outputs into a coherent next action. In project-scoped execution, the coordinator refers to the \texttt{project-main} agent in Table \ref{tab:agent_roles}; its purpose is not to execute every detail, but to keep the master project state clean enough for strategic reasoning, documentation, and human approval.

The coordinator performs a routing decision
\begin{equation}
\pi(a \mid s_t, o_t, m_t),
\end{equation}
where $s_t$ denotes the observable interaction state, $o_t$ denotes the current objective and constraint state, and $m_t$ denotes the retrieved memory state. This formulation emphasizes that delegation is conditioned not only on the latest user request, but also on accumulated operational history and the currently active scientific objective. Role specialization reduces cross-task interference and allows different agents to maintain narrower but deeper competence surfaces.

The runtime further distinguishes between shared coordination and project-scoped coordination. Shared coordination is used for global or personal interactions that do not require project-private tools, workspaces, or credentials. Project-scoped coordination is used when local project environments, project memory, and project-specific constraints must all be exposed to the agent society. This separation preserves security boundaries while reusing a common agent kernel. Table~\ref{tab:agent_roles} summarizes the canonical agent roles used in this hierarchy.

\begin{table}
    \centering
    \small
    \caption{Canonical agent roles in SCION.}
    \label{tab:agent_roles}
    \begin{tabular}{p{2.2cm}p{1.8cm}p{2.5cm}p{6.2cm}}
        \toprule
        Runtime role & Agent kind & Scope & Primary function and realization \\
        \midrule
        \texttt{shared-main} & \texttt{main} & shared & Entry coordinator for personal or non-project sessions; maintains user-level continuity and delegates only to shared-safe specialists. \\
        \texttt{project-main} & \texttt{main} & project & Clean project-level master coordinator; retains global objectives, key decisions, branch structure, approved summaries, project memory, tools, and constraints; integrates specialist outputs into project-level action. \\
        \texttt{sci} & \texttt{sci} & project specialist & Performs scientific reasoning, hypothesis refinement, experiment design, and domain-constrained analytical synthesis. \\
        \texttt{idea} & \texttt{idea} & shared specialist & Generates and judges ideas based on the scientific field and literatures found. \\
        \texttt{reading} & \texttt{reading} & shared specialist & Retrieves, reads, and synthesizes literature or evidence; optimized for grounding and summarization rather than environmental actuation. \\
        \texttt{exec} & \texttt{exec} & project specialist & Operates over commands, tools, local environments, and external compute; produces tool traces, artifacts, and runtime-side operational state. \\
        \texttt{lark} & \texttt{lark} & shared specialist & Handles collaboration-platform interaction, message synchronization, and external communication workflows. \\
        \bottomrule
    \end{tabular}
\end{table}

This taxonomy should be understood as a cognitive division of labor rather than as a software packaging choice. The distinction between coordinator agents and specialist agents allows the system to separate intent interpretation from evidence synthesis, scientific reasoning, execution, and external communication. As a result, no single agent must simultaneously optimize for all roles, and the coordinator can remain relatively stable even when specialist tools or skills evolve.

\subsection{Profile-Driven Agent Realization}
Although the agents above appear behaviorally distinct, they are all realized through the same kernel. Formally, each agent can be described by a profile
\begin{equation}
P_a = \langle \mathrm{id}_a, \kappa_a, \delta_a, \mathcal{T}_a, \mathcal{S}_a, \mathcal{M}_a, \rho_a \rangle,
\end{equation}
where $\mathrm{id}_a$ is the runtime identifier, $\kappa_a$ the agent kind, $\delta_a$ the deployment scope, $\mathcal{T}_a$ the tool profile, $\mathcal{S}_a$ the skill profile, $\mathcal{M}_a$ the memory profile, and $\rho_a$ the runtime policy. The common kernel is responsible for message normalization, context rendering, tool-loop execution, delegation handling, and result normalization. What changes across agents is therefore not the existence of a separate runtime skeleton, but the profile that governs what an agent may see, invoke, remember, and write.

This design has two architectural consequences. First, the architecture remains extensible: introducing a new specialist does not require cloning a new agent runtime, but defining a new profile over the same kernel. Second, the architecture decouples cognitive topology from deployment topology. An agent may be invoked locally, on a remote runtime, or within a project-specific execution environment, while preserving the same abstract coordination semantics from the perspective of the coordinator.

\subsection{Context Construction as a Runtime Primitive}
In SCION, context is not a passive transcript but an actively assembled execution surface. An agent never receives the entire global system state. Instead, the runtime constructs an agent-specific context
\begin{equation}
\mathcal{C}_a^{(t)} = \mathcal{I}_a \cup \mathcal{O}_t \cup \mathcal{E}_a^{(t)} \cup \mathcal{T}_a \cup \mathcal{S}_a \cup \mathcal{M}_a^{(t)},
\end{equation}
where $\mathcal{I}_a$ denotes the identity and mission specification of agent $a$, $\mathcal{O}_t$ the current objective slice and active constraints, $\mathcal{E}_a^{(t)}$ the execution environment and workspace state, $\mathcal{T}_a$ the tool surface, $\mathcal{S}_a$ the injected skill substrate, and $\mathcal{M}_a^{(t)}$ the retrieved memory. The design principle is selective exposure rather than maximal exposure: each agent receives only the information necessary for its role, thereby reducing prompt competition and limiting irrelevant state contamination.

Context construction also serves as a governance mechanism. Shared agents do not receive project-private workspaces or project-specific credentials. Project agents may access local tools and custom skills, but only within the boundary of the active project environment. In this sense, context design determines not only what an agent can reason about, but also what it is operationally permitted to do.

Before prompt rendering, the runtime first assembles a typed intermediate context representation and only then serializes it into model-facing instructions. This means that prompt generation is downstream of context modeling rather than the other way around. The architecture therefore treats context assembly as a first-class control surface. Table~\ref{tab:context_blocks} lists the main context blocks and their architectural functions.

\begin{table}
    \centering
    \small
    \caption{Structured context blocks injected into an agent.}
    \label{tab:context_blocks}
    \begin{tabular}{p{2.5cm}p{4cm}p{6.3cm}}
        \toprule
        Context block & Typical content & Architectural function \\
        \midrule
        Identity block & agent identity, role, mission, coordination duty & Establishes the cognitive posture from which the agent interprets the task. \\
        Objective block & current goal, delegation reason, active constraints, expected output & Binds reasoning to the immediate scientific objective rather than the full global history. \\
        Environment block & workspace state, runtime mode, project boundary, available resources & Defines what resources are operationally visible and actionable. \\
        Tool block & tool summaries, invocation affordances, safety hints & Exposes executable action surfaces without overwhelming the agent with raw implementation detail. \\
        Skill block & skill summaries and optional file references & Injects procedural domain knowledge while deferring full skill materialization until needed. \\
        Memory block & L1 active memory, L2 durable memory, L3 reference memory, priority policy & Restores continuity and retrieval order while preserving relevance. \\
        \bottomrule
    \end{tabular}
\end{table}

Because different agents receive different context surfaces, the system does not merely instantiate multiple copies of one assistant. It creates multiple cognitive viewpoints over the same underlying scientific process. A coordinator sees project- or user-level continuity; a reading agent sees evidentiary material; an execution agent sees the action environment; a scientific reasoning agent sees hypotheses, constraints, and domain goals. This selective asymmetry is a central property of the architecture.

\subsection{Layered Memory and Continuity Preservation}
The memory subsystem is designed to preserve continuity without flattening all history into the prompt. The system distinguishes multiple scopes of memory, including session memory, agent-private working memory, user-profile memory, project memory, shared runtime memory, artifact memory, and audit memory. For retrieval, these scopes are organized into a layered structure
\begin{equation}
\mathcal{M}_a^{(t)} = \left(M_{a,\mathrm{L1}}^{(t)}, M_{a,\mathrm{L2}}^{(t)}, M_{a,\mathrm{L3}}^{(t)}\right),
\end{equation}
where L1 captures active session continuity, L2 stores durable user and project knowledge, and L3 retains artifact- and audit-oriented reference memory. The retrieval policy prioritizes L1 over L2 and L2 over L3, while still allowing fresh user instructions to override stale records.

L1 supports short-horizon reasoning by preserving recent dialogue, task-local tool outcomes, and structured working todos. L2 records longer-lived facts such as user preferences, project decisions, scientific assumptions, and reusable operational constraints. L3 stores traces, artifacts, failure branches, and provenance records that are essential for recoverability and auditability but need not always enter the active reasoning window. This separation prevents transient local state from being mistaken for durable knowledge.

To prevent unbounded prompt growth, the system performs rolling summarization over unsummarized interaction spans and promotes sufficiently stable facts into durable memory. The result is a memory architecture that compresses experience without erasing it: recent events remain explicit, while older but still valuable information is transformed into typed summaries and long-term records. Memory access is also role-dependent, so different agents read and write different slices of the epistemic state according to their memory profiles. Table~\ref{tab:memory_scopes} summarizes the memory scopes, retrieval layers, and access semantics.

\begin{table}
    \centering
    \small
    \caption{Memory scopes, retrieval layers, and access semantics.}
    \label{tab:memory_scopes}
    \begin{tabular}{@{}L{3.4cm}C{0.7cm}L{3.1cm}L{5.2cm}@{}}
        \toprule
        Memory scope & Layer & Semantic role & Typical access discipline \\
        \midrule
        \texttt{session\_memory}
        & L1
        & Recent interaction continuity
        & Read during active dialogue; cleared with session reset. \\

        \makecell[tl]{\texttt{agent\_private\_}\\\texttt{working\_memory}}
        & L1
        & Transient todos and local plans
        & Writable by the active agent; not treated as durable knowledge unless summarized or promoted. \\

        \texttt{user\_profile\_memory}
        & L2
        & Stable user preferences and constraints
        & Updated only when facts are sufficiently persistent; primarily read by coordinator agents. \\

        \texttt{project\_memory}
        & L2
        & Project decisions, assumptions, and reusable findings
        & Available in project-scoped execution; isolated from shared coordination. \\

        \makecell[tl]{\texttt{shared\_project\_}\\\texttt{runtime\_memory}}
        & L2
        & Shared runtime-side operational state
        & Written by runtime-facing specialists when shared project execution state must persist across turns. \\

        \texttt{artifact\_memory}
        & L3
        & Generated outputs and reusable artifacts
        & Retrieved as reference evidence when prior artifacts become relevant to current action. \\

        \texttt{audit\_memory}
        & L3
        & Traces, provenance, failure records, execution evidence
        & Used for recoverability, verification, and post hoc analysis rather than routine prompt injection. \\
        \bottomrule
    \end{tabular}
\end{table}

Equally important is the distinction between working memory and durable memory. Working memory is instrumentally useful for advancing the current task but may be invalidated within a short horizon. Durable memory must survive across sessions and remain meaningful when recalled later. The architecture therefore does not simply store everything that happened; it continuously decides what should remain transient, what should be summarized, and what should become reusable knowledge.

\subsection{Governed Delegation and Long-Horizon Recovery}
Scientific workloads are often asynchronous, tool-heavy, and failure-prone. Delegation in SCION is therefore implemented as a governed runtime operation rather than as an informal conversational transfer. Each delegation carries an explicit target agent, a rationale, provenance metadata, and an execution expectation, typically ranging from immediate execution to long-running jobs. This allows the system to maintain coherence across simulation, retrieval, environment preparation, and other subtasks that cannot be completed within a single synchronous turn.

The runtime further incorporates critic checkpoints and anomaly interception mechanisms. Intermediate outputs can be evaluated against workflow consistency conditions, tool-interface constraints, domain rules, and project-specific safety boundaries. When an agent drifts from the objective, hallucinates unsupported content, or enters an unproductive branch, the runtime can trigger rollback, branch termination, re-routing, or resource reallocation. In this sense, the architecture does not merely parallelize reasoning; it governs a recoverable search process over scientific action trajectories.

The delegation layer also distinguishes user-visible outputs from internal-only intermediate states. Some specialist results are not final answers but preparatory actions, authentication checks, environment setup outcomes, or callback confirmations. These intermediate results are reintegrated into the coordinator rather than exposed directly to the scientist. Such a distinction is critical in multi-agent settings because it prevents low-level control traffic from polluting the human-facing interaction surface. Table~\ref{tab:delegation_semantics} summarizes the execution horizons supported by this delegation mechanism.

\begin{table}
    \centering
    \small
    \caption{Delegation semantics across different execution horizons.}
    \label{tab:delegation_semantics}
    \begin{tabular}{p{1.6cm}p{3.1cm}p{2.6cm}p{4.8cm}}
        \toprule
        Class & Typical tasks & Completion path & Architectural significance \\
        \midrule
        instant & clarification, bounded reasoning, lightweight retrieval & same-turn return & Preserves conversational responsiveness while still allowing role specialization. \\
        short & focused synthesis, constrained tool use, local execution loops & near-term return to coordinator & Enables localized depth without forcing the coordinator to absorb specialist reasoning directly. \\
        long & simulation, environment setup, extensive computation, callback-driven tasks & asynchronous job completion and reintegration & Supports coherent progress over tasks that exceed a single conversational or execution window. \\
        \bottomrule
    \end{tabular}
\end{table}

Taken together, these mechanisms show that SCION is not merely a prompt-engineered assistant with more tools. It is a profile-governed, memory-bearing, delegation-capable multi-agent runtime in which scientific work is distributed across specialized cognitive surfaces while remaining coordinated, inspectable, and recoverable.

\section{Scientific Discovery as Meta-Harness}
\label{sec:meta-harness}

As implemented by SCION, the Science Agent's meta-harness behavior is organized around three persistent layers. The first layer is task representation: high-level scientific intent is compiled into a Research Execution Plan that specifies objectives, constraints, dependencies, verification checkpoints, expected artifacts, and fallback conditions. The second layer is runtime execution: specialized agents, tools, and project-scoped environments are coordinated to carry out the plan while monitoring intermediate results. The third layer is memory accumulation: successful trajectories, failed attempts, artifacts, provenance, and reusable knowledge are preserved so that later decisions do not start from an empty state, as shown in Figure \ref{fig:inversesearch}.

This Science-Agent-centered organization changes the form of scientific work in four ways. First, implicit intent becomes explicit executable plans through REPs. Second, manual cross-tool coordination becomes governed multi-agent execution in project-scoped runtimes. Third, isolated intermediate outputs become persistent epistemic assets through long-term memory and retrieval. Fourth, evaluation expands from single-model accuracy to system-level properties such as recoverability, auditability, and knowledge reuse. The result is not a replacement of scientists, but a reallocation of routine coordination work toward an auditable agentic runtime, while human scientists remain responsible for problem formulation, constraints, interpretation, and final scientific judgment.

We further describe how the Science Agent in SCION uses the meta-harness formulation in two scientific discovery scenarios.

\subsection{Scientific Discovery with Known Target Property: Target-Conditioned Inverse Search}
\begin{figure}
    \centering
    \includegraphics[width=0.8\linewidth]{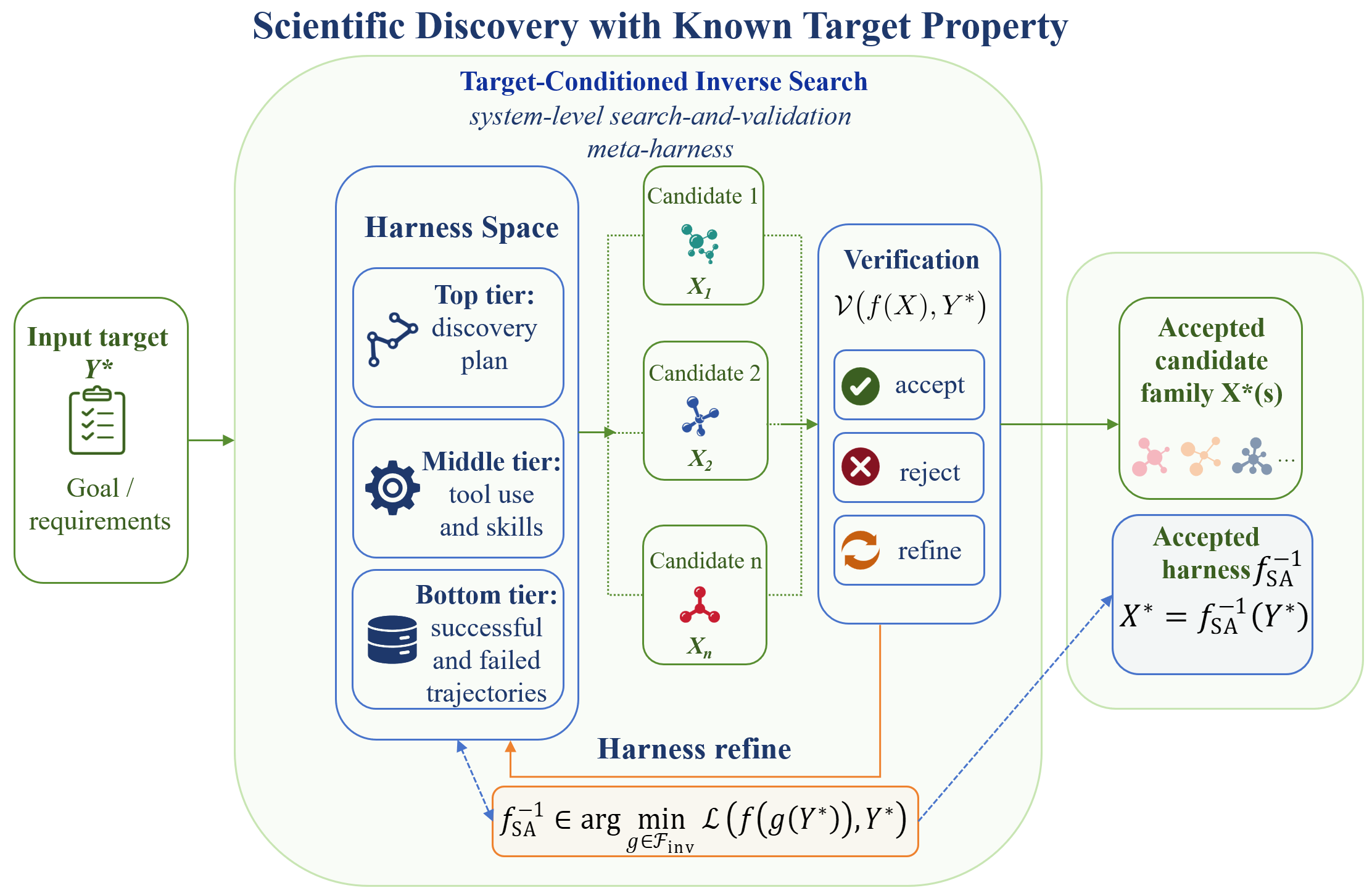}
    \caption{Science Agent for target-conditioned inverse search. A desired scientific target is transformed into executable search actions through planning, candidate generation, tool use, verification, and memory update. Under this view, Science Agent acts as an approximate inverse operator that repeatedly refines candidates rather than merely applying a forward predictive model.}
    \label{fig:inversesearch}
\end{figure}
Figure~\ref{fig:inversesearch} illustrates how the Science Agent operationalizes target-conditioned inverse search. The above shift in production relations can also be described as a shift in the mathematical form of scientific work. Let
\begin{equation}
f : \mathcal{X} \rightarrow \mathcal{Y}
\end{equation}
denote the forward scientific process that maps a candidate object \(X \in \mathcal{X}\) to an observed or validated outcome \(Y \in \mathcal{Y}\), such that
\begin{equation}
Y = f(X).
\end{equation}
Depending on the domain, \(X\) may denote a molecular structure, a protein sequence, a material configuration, a spectral assignment, or a synthesis protocol, while \(Y\) may denote the corresponding functional property profile, structural behavior, experimental response, or task-specific performance target.

In practical scientific discovery, the objective is usually not to predict \(Y\) from a known \(X\), but to identify an \(X^*\) that satisfies a desired target \(Y^*\). This can be expressed as a target-conditioned inverse problem:
\begin{equation}
f(X^*) \approx Y^*.
\end{equation}
In the ideal case, one would solve
\begin{equation}
X^* = f^{-1}(Y^*).
\end{equation}
Here, \(f^{-1}\) denotes an ideal inverse map. However, in real scientific settings the inverse \(f^{-1}\) is typically non-unique, partially observable, computationally intractable, or not explicitly available. Therefore, the practical objective is not to recover a closed-form global inverse, but to search for feasible candidates whose validated outcomes approach the target:
\begin{equation}
X^* \in \arg\min_{X \in \mathcal{X}} \mathcal{L}\bigl(f(X), Y^*\bigr),
\end{equation}
where \(\mathcal{L}(\cdot,\cdot)\) should be understood operationally not merely as an abstract discrepancy measure, but as the output of a verifier component that tests whether the realized outcome \(f(X)\) satisfies the requirements encoded in \(Y^*\). In practice, this verifier may combine threshold checks, rule-based constraints, physical consistency tests, safety filters, and domain-specific acceptance criteria into a unified decision mechanism. Equivalently, one may write the corresponding verifier as
\begin{equation}
\mathcal{V}\bigl(f(X), Y^*\bigr),
\end{equation}
where \(\mathcal{V}\) returns either an acceptance decision or a verification score indicating whether \(f(X)\) is adequate for the target. Under this view, the search process is not only minimizing a loss, but repeatedly proposing candidates and invoking a verifier to determine whether they should be accepted, rejected, or refined.

From this perspective, the role of Science Agent is to operationalize a target-conditioned approximate inverse procedure. We denote this realized procedure informally as
\begin{equation}
X^* = f^{-1}_{\mathrm{SA}}(Y^*).
\end{equation}
Here, \(f^{-1}_{\mathrm{SA}}\) should not be interpreted as an analytic inverse function comparable to the ideal \(f^{-1}\). Instead, it denotes the inverse-search procedure induced by the Science Agent runtime, where the subscript \(\mathrm{SA}\) marks that the inverse is realized by the Science Agent rather than computed analytically. Given a target \(Y^*\), the system first compiles the target and its constraints into a Research Execution Plan, then generates and evaluates candidate objects through coordinated agents and tools, verifies intermediate outcomes, and updates memory so that later search steps are conditioned on prior successes and failures.

At the system level, this procedure can be viewed as selecting an approximate inverse-search strategy from a feasible procedure class:
\begin{equation}
f^{-1}_{\mathrm{SA}}\in \arg\min_{g \in \mathcal{F}_{\mathrm{inv}}} \mathcal{L}\bigl(f(g(Y^*)), Y^*\bigr).
\end{equation}
In this expression, \(\mathcal{F}_{\mathrm{inv}}\) denotes the class of inverse-search procedures that can be composed by the system under the available resources, tools, agents, and constraints. A procedure \(g \in \mathcal{F}_{\mathrm{inv}}\) may involve different choices of planning strategy, candidate-generation method, simulation or prediction toolchain, verification rule, rollback condition, and memory-update policy. Thus, the optimization is not over closed-form inverse functions, but over executable scientific search procedures. SCION does not solve this selection in closed form; rather, it approximates it online, compiling an initial procedure from the Research Execution Plan and then revising the active choices of generation, tool use, verification, and rollback as intermediate evidence accumulates.

More concretely, \(f^{-1}_{\mathrm{SA}}\) can be understood schematically as a composition of runtime components:
\[
f^{-1}_{\mathrm{SA}}
\;\approx\;
\mathrm{MemoryUpdate}
\circ
\mathrm{Verify}
\circ
\mathrm{Execute}
\circ
\mathrm{Generate}
\circ
\mathrm{Plan}.
\]
Here, \(\circ\) denotes functional composition, with the execution order proceeding from right to left. Given a target \(Y^*\), the system first plans, then generates candidate objects, executes evaluations through tools or simulations, verifies the resulting evidence, and finally updates memory for subsequent search steps.
The planner translates \(Y^*\) and its constraints into a REP. The generator proposes candidate objects or search branches. The executor invokes models, simulations, databases, or experimental interfaces to evaluate candidates. The verifier determines whether the observed or predicted outcome satisfies the target requirements. The memory updater records successful, failed, and ambiguous trajectories so that subsequent search is conditioned on accumulated evidence. This decomposition is schematic rather than a fixed pipeline: different scientific domains may instantiate these components with different agents, tools, and verification criteria. Concretely, these stages map onto SCION's runtime: planning corresponds to REP compilation by the coordinator, generation primarily to the \texttt{idea} and \texttt{sci} agents, execution to the \texttt{exec} agent and external tools, verification to critic checkpoints and domain verifiers, and memory update to the layered memory subsystem. Section~\ref{sec:experiments} evaluates two such instantiations: multi-property molecule generation as target-conditioned inverse search, and target-specific antibody screening as its hidden-target counterpart.

This formulation clarifies why SCION is more than a collection of independent AI tools. A conventional pipeline applies isolated predictors to estimate fragments of the forward map \(f\). By contrast, SCION attempts to construct and revise an executable inverse-search process associated with \(Y^*\), under explicit constraints, intermediate verification, rollback conditions, and memory reuse. The essence of the system is therefore not only better prediction, but the conversion of scientific discovery into a governed procedure for approximate inverse design.

\subsection{Scientific Discovery with Hidden Target Property: Batch Active Search}
\label{sec:batch_active_search}
\begin{figure}
    \centering
    \includegraphics[width=0.8\linewidth]{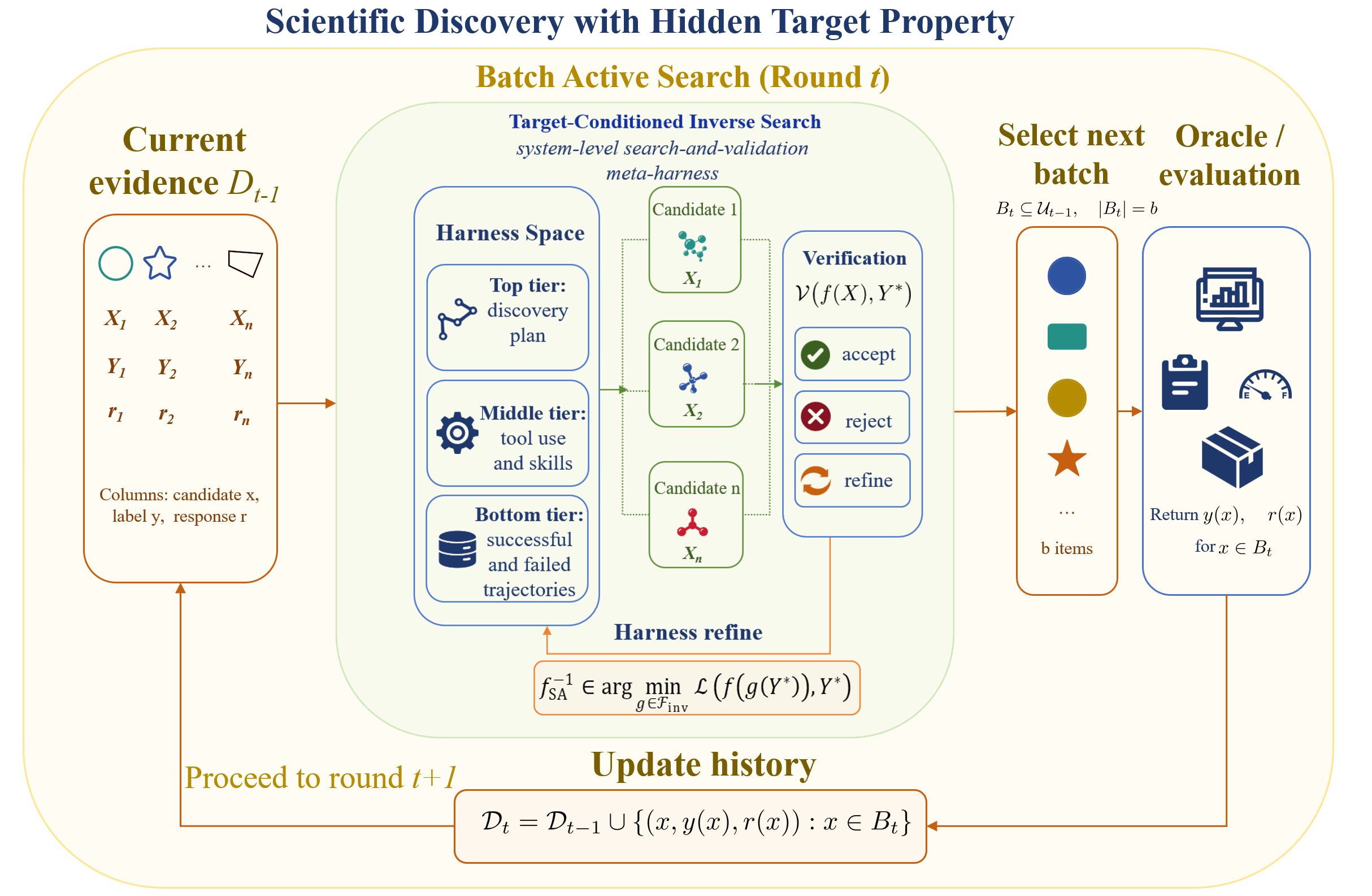}
    \caption{SCION for batch active search under hidden target properties. Each round selects candidate batches, receives limited feedback, updates the current belief state, and revises the next-round search policy. This setting captures discovery problems where the target criterion is only partially revealed through costly observations.}
    \label{fig:batchactivesearch}
\end{figure}
In many scientific discovery settings, the specified target property $Y^*$ is not explicitly known to the researchers. A verifier may be available, but the underlying property that makes a candidate desirable may only be revealed through resource-limited observations. Such a problem can be modeled as target-conditioned inverse search with a hidden target property, which naturally leads to a batch active search formulation shown in Figure~\ref{fig:batchactivesearch}.

We consider a pool-based batch active search problem over a finite candidate set
\(
\mathcal{X} = \{x_i\}_{i=1}^{N}
\).
Each candidate \(x_i\) is associated with an unknown scalar property
\(
r_i \in \mathbb{R}
\),
which measures the scientific quantity of interest.
A candidate is regarded as a target if its property exceeds a predefined threshold \(\tau\), inducing a binary label
\begin{equation}
y_i = \mathbf{1}[r_i \ge \tau], \qquad y_i \in \{0,1\}.
\end{equation}
Thus, the binary label is thresholded from the underlying property value.

At round \(t \in \{1,\dots,H\}\), the learner selects a batch
\(
B_t \subseteq \mathcal{U}_{t-1}
\)
of size
\(
|B_t| = b
\),
where \(\mathcal{U}_{t-1}\) denotes the set of unlabeled candidates remaining after round \(t-1\).
The oracle then returns, for each queried candidate \(x \in B_t\), both its binary label \(y(x)\) and its underlying property value \(r(x)\).
The observation history is updated as
\begin{equation}
\mathcal{D}_t
=
\mathcal{D}_{t-1}
\cup
\{(x, y(x), r(x)) : x \in B_t\}.
\end{equation}
The total query budget is \(T = Hb\), where \(H\) is the number of experimental rounds and \(b\) is the batch size.

Following the standard active search formulation, the utility of an observed set
\(\mathcal{D}\) is defined as the number of targets discovered:
\begin{equation}
u(\mathcal{D}) = \sum_{(x,y,r)\in\mathcal{D}} y.
\end{equation}
The goal of batch active search is therefore to design a policy
\(
\pi = \{\pi_t\}_{t=1}^{H}
\)
that sequentially selects batches so as to maximize the expected number of positives found by the end of the budget:
\begin{equation}
\max_{\pi}\;
\mathbb{E}_{\pi}\!\left[u(\mathcal{D}_H)\right].
\end{equation}
Equivalently, at each round \(t\), the Bayesian optimal batch decision can be written as
\begin{equation}
B_t^\star
=
\arg\max_{\substack{B \subseteq \mathcal{U}_{t-1}\\ |B|=b}}
\mathbb{E}\!\left[
u(\mathcal{D}_H)
\mid
\mathcal{D}_{t-1}, B
\right].
\end{equation}
This objective is nonmyopic, since the value of the current batch depends not only on the immediate positives it contains, but also on how the newly revealed observations reshape posterior beliefs and downstream batch selections. Computing \(B_t^\star\) exactly is intractable, since it requires marginalizing over all future observations and batch choices. SCION therefore does not evaluate this expectation directly; instead, its Science Agent approximates the round-\(t\) policy from the accumulated history \(\mathcal{D}_{t-1}\), as formalized below.

Let
\begin{equation}
p_{t-1}(x)
:=
\Pr\!\left(y(x)=1 \mid \mathcal{D}_{t-1}\right)
=
\Pr\!\left(r(x)\ge\tau \mid \mathcal{D}_{t-1}\right)
\end{equation}
denote the posterior probability that candidate \(x\) is positive given the current history.
A myopic batch strategy selects the batch with the largest immediate expected gain,
\begin{equation}
B_t^{\text{myopic}}
=
\arg\max_{\substack{B \subseteq \mathcal{U}_{t-1}\\ |B|=b}}
\sum_{x\in B} p_{t-1}(x),
\end{equation}
which ignores the effect of the current observations on future rounds.

Importantly, although the active search objective is defined over binary target labels, the returned property values provide additional graded information beyond the thresholded feedback.
This is particularly useful in sparse-reward regimes where queried batches may contain few or even no positive labels, i.e., \(y(x)=0\) for most or all queried candidates.
In such cases, binary observations alone provide only weak signal for policy improvement, whereas the continuous property values \(r(x)\) still induce relative ranking information over candidates and support a coarse surrogate characterization of the local search landscape.
The role of the numerical feedback is therefore not to replace the target label, but to densify the learning signal under thresholded verification, so that subsequent batch decisions can be guided even when the observed labels are uniformly negative.

This formulation also clarifies the relation between batch active search and the target-conditioned inverse search introduced in the previous subsection.
There, scientific discovery was modeled as the problem of finding a feasible approximate inverse
\(
f^{-1}_{\mathrm{SA}}(Y^*)
\)
under verifier-mediated feedback.
With hidden target property, however, this inverse search cannot be executed as an unconstrained global procedure.
Instead, it is unfolded into a sequential decision process, in which the system repeatedly allocates limited queries, observes feedback, and revises its search policy over rounds.
In this sense, batch active search provides a budget-constrained operational formulation of scientific discovery under hidden targets.

More specifically, at round \(t\), once the system has received the accumulated history
\(
\mathcal{D}_{t-1}
\),
the problem is no longer simply to estimate which candidates are positive in isolation.
Rather, the problem is to use the current evidence to optimize the next-round search action in a target-conditioned manner.
That is, each round instantiates a conditional inverse-search problem of the form
\begin{equation}
\pi_t
\approx
f^{-1}_{\mathrm{SA}}\!\left(\cdot \mid \mathcal{D}_{t-1}\right),
\end{equation}
where the $Y^*$ is hidden in $\mathcal{D}_{t-1}$ and the output is not a final solution candidate \(X^*\) in one step, but a round-specific batch-selection strategy for choosing \(B_t\).
The feedback obtained from previous rounds, including both the binary verifier outputs and the underlying property values, therefore serves to reshape the approximate inverse operator used at the current round.
Equivalently, each round may be viewed as solving a local target-conditioned inverse search problem whose purpose is to improve the single-round policy under the remaining budget.

From this perspective, the role of \textsc{SCION} is to operationalize this iterative policy-level inverse search.
After each experimental round, \textsc{SCION} takes the updated history
\(
\mathcal{D}_{t-1}
\)
as input, interprets the observed success and failure patterns, exploits the graded property feedback to recover coarse ordering structure even under all-negative labels, and produces a revised strategy for the next batch.
What is being optimized is therefore not only a static predictor over \(\mathcal{X}\), but the history-conditioned search procedure itself.
In this sense, \textsc{SCION} functions as an adaptive inverse operator over policies: it converts round-by-round experimental feedback into improved single-round decision rules, so that the overall multi-round process progressively approximates a budget-constrained target-conditioned inverse search. Section~\ref{sec:experiments} evaluates exactly this single-round policy: given the screening history \(\mathcal{D}_{t-1}\), each system must select the next batch, and performance is measured by how well the selected candidates recover the true high-value set.

\section{Scientific Applications}

This section illustrates how the two discovery formulations above can be instantiated in representative scientific domains. The purpose is not to re-derive the inverse-search or active-search objectives, but to show how their variables, feedback signals, constraints, and executable procedures correspond to concrete research workflows. Table~\ref{tab:application_mapping} summarizes the mapping across three representative applications: theoretical materials analysis, multi-property molecular design, and target-specific antibody screening.

\begin{table}[!htbp]
    \centering
    \small
    \caption{Application-level instantiations of the SCION discovery formulations.}
    \label{tab:application_mapping}
    \begin{tabularx}{\textwidth}{@{}P{2.7cm}P{3.0cm}Y Y@{}}
        \toprule
        Application
        & Candidate space
        & Target, feedback, and constraints
        & SCION role \\
        \midrule

        Theoretical materials analysis
        & Compositions, crystal structures, doped configurations, defects, surfaces
        & Target material properties such as stability, band gap, adsorption energy, or transport behavior; constrained by physical plausibility, charge balance, thermodynamic stability, and simulation consistency.
        & Coordinates literature grounding, structure preparation, simulation scheduling, post-processing, verification, and mechanistic synthesis. \\

        \midrule

        Multi-property molecular design
        & Molecular graphs, SMILES strings, scaffolds, scaffold derivatives
        & Target molecular properties such as potency, solubility, permeability, toxicity risk, ADMET, or synthesizability; constrained by chemical validity, synthesizability, safety filters, and multi-objective trade-offs.
        & Coordinates candidate generation, property prediction, docking or simulation, retrosynthetic assessment, filtering, ranking, and memory reuse. \\

        \midrule

        Target-specific antibody screening
        & Antibody sequences, nanobodies, protein binders, scaffold-conditioned candidates
        & Binary hit labels and numerical feedback such as affinity, enrichment, expression yield, or developability score; constrained by limited screening budget, rare positives, specificity, stability, and manufacturability.
        & Uses historical feedback to select the next batch, update beliefs over the pool, and improve budget-constrained screening decisions. \\

        \bottomrule
    \end{tabularx}
\end{table}

\FloatBarrier

\subsection{Theoretical Analysis of Materials}

Theoretical materials analysis instantiates the target-conditioned inverse-search setting. A candidate \(X \in \mathcal{X}_{\mathrm{mat}}\) may denote a composition, crystal structure, doped configuration, defect configuration, or surface arrangement \cite{CURTAROLO2012218,10.1063/1.4812323}. The forward process maps such a material candidate to theoretical or computational observables:
\begin{equation}
f_{\mathrm{mat}} : \mathcal{X}_{\mathrm{mat}} \rightarrow \mathcal{Y}_{\mathrm{mat}},
\end{equation}
where \(\mathcal{Y}_{\mathrm{mat}}\) may include formation energy, band gap, magnetic response, adsorption energy, stability metrics, transport coefficients, or mechanistic descriptors \cite{xie2018crystal,Chen_2019,Chen_2022_periodic,Ryczko_2019}. The discovery objective is to identify feasible material candidates whose validated responses approach a target property profile \(Y^*\).

In this domain, the main difficulty is not only prediction, but the operational coordination of a long theoretical workflow. Candidate structures must be generated or retrieved, simulation inputs must be prepared, computational jobs must be scheduled, numerical outputs must be post-processed, and physically implausible or inconsistent results must be detected. The feasible set \(\mathcal{X}_{\mathrm{feasible}}\) may impose charge balance, structural stability, thermodynamic plausibility, known synthesis boundaries, or project-specific physical constraints.

SCION realizes the corresponding approximate inverse-search process by compiling the target material profile and constraints into a Research Execution Plan. The plan may include stages for literature grounding, structure generation, simulation, post-processing, anomaly detection, and mechanistic synthesis. Specialized agents can then coordinate computation, evidence retrieval, and analysis across multiple branches, while verifier components check physical consistency and trigger rollback or refinement when outputs violate domain constraints. The resulting traces, failed structures, discarded hypotheses, and successful candidates are stored as reusable memory for subsequent materials investigations.

\subsection{Multi-Property Molecular Design}

Multi-property molecular design provides another instance of target-conditioned inverse search, but with a different candidate space and constraint structure. Here, \(X \in \mathcal{X}_{\mathrm{mol}}\) denotes a candidate molecular structure, represented for example as a graph, a SMILES (Simplified Molecular-Input Line-Entry System) string, scaffold, or scaffold derivative \cite{doi:10.1021/acs.jcim.5c02234,tang2024surveygenerativeainovo,jin2019junctiontreevariationalautoencoder,olivecrona2017molecularnovodesigndeep}. The forward process evaluates a molecule into a vector of relevant properties:
\begin{equation}
f_{\mathrm{prop}}(X)
=
\bigl(y_1(X), y_2(X), \dots, y_m(X)\bigr),
\end{equation}
where the properties may include potency, selectivity, solubility, permeability, metabolic stability, toxicity risk, ADMET (absorption, distribution, metabolism, excretion, and toxicity) profiles, and synthesizability \cite{Bickerton_2012}. The goal is to find molecules whose property vectors approach a target profile \(Y^*\) while satisfying hard chemical and project-specific constraints.

Compared with materials analysis, molecular design is characterized by a larger discrete search space and stronger validity constraints. A generated candidate must be chemically valid, preferably synthesizable, safe under predefined filters, and competitive under multiple objectives. In practice, the output is often not a single optimum but a ranked candidate set or an approximate Pareto frontier reflecting trade-offs among desired properties.

SCION supports this workflow by coordinating candidate generation, property evaluation, retrosynthetic or synthesizability assessment, docking or simulation, toxicity filtering, and cross-property ranking. Verifier components remove invalid SMILES, infeasible structures, candidates that violate hard filters, or molecules that drift away from the target profile. Memory is especially useful in this setting because rejected motifs, failed optimization paths, and previously observed trade-off patterns can be reused in later design rounds rather than rediscovered from scratch.

\subsection{Target-Specific Antibody Screening}

Target-specific antibody screening instantiates the hidden-target batch active-search setting. Given a target protein or antigen \(T^*\), the system operates over a finite candidate pool
\begin{equation}
\mathcal{X} = \{x_i\}_{i=1}^{N},
\end{equation}
where each \(x_i\) may represent an antibody sequence, nanobody, protein binder, scaffold-conditioned design, or another target-specific biomolecular construct \cite{doi:10.1126/science.add2187,kyro2025modelcentricreviewdeeplearning}. Unlike the previous two applications, the target criterion is often only partially revealed through costly evaluations such as binding assays, enrichment experiments, affinity prediction, structure-based screening, expression tests, or developability assessment \cite{cheng2024alphafolding4ddiffusiondynamic,doi:10.1126/science.adg7492}.

For each queried candidate, the oracle may return both a binary hit decision and a numerical feedback value. The binary label indicates whether the candidate satisfies a task-specific acceptance criterion, such as sufficient binding affinity, target specificity, stability, manufacturability, or developability. The numerical value may correspond to binding affinity, inhibition potency, enrichment score, expression yield, developability score, or a scalarized screening metric. This numerical feedback is important because early screening rounds may contain few positive hits, while graded scores can still provide relative ranking information among negative or near-miss candidates.

SCION supports this setting by converting historical screening feedback into a next-round batch-selection strategy. After each round, the system updates its belief over the candidate pool, interprets positive, negative, and near-threshold observations, and selects a new batch under the remaining budget. In this case, the role of SCION is not to search for a single final candidate in one step, but to improve the history-conditioned screening policy over multiple rounds. This makes target-specific antibody screening a natural example of budget-constrained discovery under hidden target properties.

\section{Experiments}
\label{sec:experiments}
This section evaluates whether SCION improves system-level scientific work beyond isolated model answering. The evaluation covers five tasks, including four single-step tasks: scientific reading and reasoning in physics, idea generation, multi-property molecular generation, and target-specific antibody screening, and one end-to-end test on ResearchClawBench\cite{xu2026researchclawbenchbenchmarkendtoendautonomous}. We first describe the shared baselines and implementation details, then report the task setting, dataset, evaluation metrics, and results for each benchmark.

\paragraph{Baselines.}
The baselines fall into two groups. The first consists of pipeline-based autonomous research agents, including AI-Researcher \cite{tang2025airesearcherautonomousscientificinnovation}, AI-Scientist-v2 \cite{yamada2025aiscientistv2workshoplevelautomated}, InternAgent-1.0 \cite{internagentteam2025internagentagentscientist}, and AutoResearchClaw \cite{liu2026autoresearchclaw}. The second consists of agentic research systems, including EvoScientist \cite{lyu2026evoscientistmultiagentevolvingai}, DR-Claw \cite{song2026drclaw}, and ScienceClaw \cite{wang2026autonomousagentscoordinatingdistributed}. We also include ARIS~\cite{yang2026arisautonomousresearchadversarial}, a collaborative system built around Claude Code and Codex. Each experiment reports the subset of baselines available for that benchmark.

\paragraph{Implementation details.}
Within each benchmark, all systems receive the same task inputs and context constraints. In Scientific Reading, SCION and all reported baselines use Kimi-K2.5~\cite{kimi_k25_model} on the same 520 CMPhysBench questions with the same evaluation metrics. This comparison controls for backbone choice. ARIS is not included in Scientific Reading.

For Idea Maker, Molecule Generator, and Antibody Screening, SCION uses a locally deployed DeepSeek-V3.1-Nex-N1.1 model~\cite{nex,nex_n11_model}, abbreviated as Nex-N1.1. Its publicly available weights are released under Apache-2.0. The model is post-trained from the DeepSeek-V3.1-Base mixture-of-experts backbone, which has 671B total parameters and approximately 37B activated per token~\cite{deepseek_v31_base}. The identifiers \texttt{nex-n1.1} and \texttt{nex-n1.1-public} refer to this model in our experimental configurations.

The original ARIS baseline uses Claude Opus 4.6 in Claude Code for task execution and GPT-5.5 in Codex for critique and review, following the cross-model collaboration design of ARIS~\cite{yang2026arisautonomousresearchadversarial}. We additionally evaluate a task-adapted configuration, denoted ARIS + Nex-N1.1, in which the executor, critic, and reviewer all use DeepSeek-V3.1-Nex-N1.1, matching SCION's backbone. Section~\ref{sec:backbone-controlled-aris} reports these comparisons, which control for backbone choice. SCION is evaluated as an agentic workflow that decomposes tasks, preserves intermediate evidence, and produces normalized outputs for the corresponding evaluation protocol. For pipeline agents such as AI-Researcher, we adapt their subagents to the specified tasks.

\paragraph{Repeated-run reporting.}
For Scientific Reading, Molecule Generator, and Antibody Screening, we evaluate each method over three independent runs and report the mean and sample standard deviation of the benchmark score across runs. Each bar shows the mean, and the error bars extend one sample standard deviation above and below it. Tables~\ref{tab:reading_three_runs}--\ref{tab:antibody_three_runs} provide the complete numerical results.

\subsection{Scientific Reading}

\paragraph{Task setting.}
In Scientific Reading, each system must solve open-ended physics problems by constructing a complete reasoning path rather than selecting from predefined answers. All systems receive the same problem statement and are required to generate both a solution process and a final answer.

\paragraph{Dataset.}
We evaluate on CMPhysBench \cite{wang2025cmphysbenchbenchmarkevaluatinglarge}, a condensed-matter physics benchmark containing 520 graduate-level calculation problems. The benchmark covers representative subfields such as magnetism, superconductivity, strongly correlated systems, semiconductors, and theoretical foundations.

\paragraph{Evaluation metrics.}
The primary metrics are answer accuracy and SEED score from CMPhysBench \cite{wang2025cmphysbenchbenchmarkevaluatinglarge}. Accuracy measures whether the final answer is equivalent to the reference answer. The SEED score, introduced with CMPhysBench~\cite{wang2025cmphysbenchbenchmarkevaluatinglarge}, provides partial credit by comparing symbolic or mathematical expressions through a scalable expression-edit-distance procedure.

\paragraph{Results.}

\begin{figure}[htbp]
    \centering
\begin{tikzpicture}
    \begin{axis}[
        ybar,
        width=0.84\textwidth,
        height=6.8cm,
        bar width=5pt,
        ymin=0,
        ymax=50,
        ylabel={Score},
        xmin=0.45,
        xmax=2.43,
        xtick={1,1.88},
        xticklabels={SEED, Accuracy (\%)},
        xticklabel style={font=\small},
        legend style={at={(0.5,1.12)}, anchor=south, legend columns=4, font=\scriptsize},
        enlarge x limits=false,
        major grid style={dashed,gray!30},
        ymajorgrids=true,
        error bars/y dir=both,
        error bars/y explicit,
        error bars/error bar style={draw=black,line width=0.45pt},
        error bars/error mark options={rotate=90,mark size=1.2pt,line width=0.45pt}
    ]
        \addplot+[draw=SCIONBar, fill=SCIONBar!78] coordinates {(1,44.0245) +- (0,0.5447) (1.88,35.8974) +- (0,0.2938)};
        \addplot+[draw=DRClawBar, fill=DRClawBar!78] coordinates {(1,34.3249) +- (0,0.3194) (1.88,26.2179) +- (0,0.2938)};
        \addplot+[draw=AutoResearchBar, fill=AutoResearchBar!78] coordinates {(1,26.5146) +- (0,1.0307) (1.88,22.0513) +- (0,0.4003)};
        \addplot+[draw=ScienceClawBar, fill=ScienceClawBar!78] coordinates {(1,21.9439) +- (0,0.7718) (1.88,16.6667) +- (0,0.2938)};
        \addplot+[draw=AIResearcherBar, fill=AIResearcherBar!78] coordinates {(1,38.6594) +- (0,0.2552) (1.88,30.1282) +- (0,0.2221)};
        \addplot+[draw=AIScientistBar, fill=AIScientistBar!78] coordinates {(1,38.2822) +- (0,0.1556) (1.88,29.7436) +- (0,0.1110)};
        \addplot+[draw=EvoScientistBar, fill=EvoScientistBar!78] coordinates {(1,42.3986) +- (0,0.6228) (1.88,34.0385) +- (0,0.5769)};
        \addplot+[draw=InternAgentBar, fill=InternAgentBar!78] coordinates {(1,37.2874) +- (0,1.5638) (1.88,29.3590) +- (0,0.4840)};
        \legend{SCION, DR-Claw, AutoResearchClaw, ScienceClaw, AI-Researcher, AI-Scientist-v2, EvoScientist, InternAgent-1.0}
    \end{axis}
\end{tikzpicture}
    \caption{Metric-wise Scientific Reading results on CMPhysBench. All reported systems use Kimi-K2.5 on the same 520 questions. Bars show mean SEED score and final-answer accuracy across three independent runs; error bars show $\pm 1$ sample standard deviation. Higher values indicate better performance.}
    \label{fig:physicsbench}
\end{figure}

As shown in Figure~\ref{fig:physicsbench}, SCION achieves the highest mean performance, with a SEED score of $44.02 \pm 0.54$ and an accuracy of $(35.90 \pm 0.29)\%$. EvoScientist is the strongest baseline on both metrics, with $42.40 \pm 0.62$ SEED and $(34.04 \pm 0.58)\%$ accuracy. SCION improves the mean SEED score by 1.63 points and accuracy by 1.86 percentage points.

\subsection{Idea Maker}

\paragraph{Task setting.}
The Idea Maker benchmark evaluates structured scientific research ideation. Given a research query, each system is required to generate a research proposal that includes the motivation, literature gap, core hypothesis, proposed method, experimental validation plan, and expected scientific contribution.

\paragraph{Dataset.}
Following the idea-generation setting in EvoScientist~\cite{lyu2026evoscientistmultiagentevolvingai}, we use a test set of 30 AI research queries covering diverse contemporary topics in artificial intelligence. Each query is used to prompt SCION and each baseline system under the same task instruction and output requirements.

\paragraph{Evaluation metrics.}
We evaluate research proposals along two dimensions: novelty and clarity. Novelty measures whether a research idea is original, non-trivial, and meaningfully different from existing research directions. Clarity measures how clearly and coherently a proposal communicates its research objectives, proposed methods, and experimental plan, including the logical organization of the presentation. Reporting these dimensions separately characterizes both the originality of a research idea and how effectively it is communicated.
Following the pairwise evaluation setting in EvoScientist~\cite{lyu2026evoscientistmultiagentevolvingai}, we compare the proposals generated by SCION and each baseline for the same query and record Win, Tie, or Lose from SCION's perspective. The GPT-5.5 novelty evaluation in Table~\ref{tab:idea_maker} covers 30 queries per baseline, with each proposal pair judged in both presentation orders, yielding 60 judgments per comparison. We additionally report GPT-5.5 clarity results and evaluate the same proposal set with GLM-5.3-Flash on novelty and clarity. These supplementary results are reported as query-level counts over 30 queries per baseline; novelty win rate is the number of wins divided by the number of queries.

\paragraph{Results.}

\begin{table}[htbp]
    \centering
    \small
    \begin{tabular}{lccc}
        \toprule
        \textbf{Comparison} & \textbf{Win} & \textbf{Tie} & \textbf{Lose} \\
        \midrule
        SCION vs AI-Researcher & 88.33 & 0.00 & 11.67 \\
        SCION vs AI-Scientist-v2 & 86.67 & 0.00 & 13.33 \\
        SCION vs EvoScientist & 98.33 & 0.00 & 1.67 \\
        SCION vs InternAgent-1.0 & 88.33 & 0.00 & 11.67 \\
        SCION vs AutoResearchClaw & 100.00 & 0.00 & 0.00 \\
        SCION vs DR-Claw & 61.67 & 1.67 & 36.67 \\
        SCION vs ScienceClaw & 100.00 & 0.00 & 0.00 \\
        SCION vs ARIS & 91.67 & 1.67 & 6.67 \\
        \bottomrule
    \end{tabular}
    \caption{Pairwise novelty evaluation for the Idea Maker task using GPT-5.5 as the LLM judge. Entries are Win/Tie/Lose percentages from the perspective of SCION. Each SCION--baseline comparison is evaluated over 30 queries with swapped answer order, resulting in 60 pairwise judgments per comparison. ARIS uses the original Claude Opus 4.6 and GPT-5.5 configuration.}
    \label{tab:idea_maker}
\end{table}

Table~\ref{tab:idea_maker} shows that SCION wins the novelty comparison against every evaluated baseline by a majority margin. It reaches a 100.00\% win rate against AutoResearchClaw and ScienceClaw, and a 98.33\% win rate against EvoScientist. The closest comparison is DR-Claw, where SCION obtains a 61.67\% win rate, a 1.67\% tie rate, and a 36.67\% lose rate. These results suggest that, under GPT-5.5 pairwise judging, SCION more frequently produces ideas judged as more novel than those produced by existing autonomous research-agent baselines.

\paragraph{Clarity evaluation.}
Table~\ref{tab:idea_maker_clarity} reports GPT-5.5 clarity judgments for 30 valid queries per baseline. SCION achieves more wins than losses against five of the seven baselines, while AI-Scientist-v2 and DR-Claw perform better on clarity. The separate novelty and clarity results reveal differences in the systems' strengths in developing and communicating research ideas.

\begin{table}[htbp]
    \centering
    \small
    \begin{tabular}{lcccc}
        \toprule
        \textbf{Baseline} & \textbf{Queries} & \textbf{Win} & \textbf{Tie} & \textbf{Lose} \\
        \midrule
        AI-Researcher & 30 & 18 & 2 & 10 \\
        AI-Scientist-v2 & 30 & 11 & 1 & 18 \\
        EvoScientist & 30 & 18 & 5 & 7 \\
        InternAgent-1.0 & 30 & 20 & 5 & 5 \\
        AutoResearchClaw & 30 & 18 & 3 & 9 \\
        DR-Claw & 30 & 9 & 3 & 18 \\
        ScienceClaw & 30 & 28 & 0 & 2 \\
        \bottomrule
    \end{tabular}
    \caption{GPT-5.5 clarity evaluation for the Idea Maker task. Entries are query-level Win/Tie/Lose counts from SCION's perspective, with 30 valid queries per baseline.}
    \label{tab:idea_maker_clarity}
\end{table}

\paragraph{Evaluation with GLM-5.3-Flash.}
Table~\ref{tab:idea_maker_glm} reports the additional evaluation with GLM-5.3-Flash. SCION achieves novelty win rates of 70.00\%--100.00\% across the seven baselines. Across all seven baselines, both evaluators agree on which system achieves more wins in clarity: SCION leads against AI-Researcher, EvoScientist, InternAgent-1.0, AutoResearchClaw, and ScienceClaw, while AI-Scientist-v2 and DR-Claw lead on clarity.

\begin{table}[htbp]
    \centering
    \small
    \setlength{\tabcolsep}{5pt}
    \begin{tabular}{lcccc}
        \toprule
        \textbf{Baseline} & \textbf{Queries} & \makecell{\textbf{Novelty} \\ \textbf{Win/Tie/Lose}} & \makecell{\textbf{Novelty} \\ \textbf{win rate (\%)}} & \makecell{\textbf{Clarity} \\ \textbf{Win/Tie/Lose}} \\
        \midrule
        AI-Researcher & 30 & 29/0/1 & 96.67 & 20/1/9 \\
        AI-Scientist-v2 & 30 & 21/3/6 & 70.00 & 11/0/19 \\
        EvoScientist & 30 & 29/1/0 & 96.67 & 21/3/6 \\
        InternAgent-1.0 & 30 & 30/0/0 & 100.00 & 27/0/3 \\
        AutoResearchClaw & 30 & 30/0/0 & 100.00 & 18/3/9 \\
        DR-Claw & 30 & 22/1/7 & 73.33 & 8/2/20 \\
        ScienceClaw & 30 & 30/0/0 & 100.00 & 28/0/2 \\
        \bottomrule
    \end{tabular}
    \caption{GLM-5.3-Flash evaluation of novelty and clarity for the Idea Maker task. Win/Tie/Lose counts are reported from SCION's perspective over 30 queries per baseline. Novelty win rate is Win/30.}
    \label{tab:idea_maker_glm}
\end{table}

\paragraph{Human evaluation and agreement with GPT-5.5.}
We invited a second-year PhD student in artificial intelligence to evaluate the research proposals. For each proposal pair included in the human evaluation, the annotator read the proposals generated by SCION and the corresponding baseline and compared their novelty and clarity separately. We then compared the human annotations with the corresponding GPT-5.5 judgments, reporting observed agreement and Cohen's $\kappa$, which adjusts for chance agreement.

\begin{table}[htbp]
    \centering
    \small
    \begin{tabular}{lccc}
        \toprule
        \textbf{Dimension} & \makecell{\textbf{Human--GPT-5.5} \\ \textbf{agreement (\%)}} & \textbf{Cohen's $\kappa$} & \textbf{95\% CI for $\kappa$} \\
        \midrule
        Novelty & 83.94 & 0.472 & 0.302--0.624 \\
        Clarity & 82.38 & 0.690 & 0.594--0.780 \\
        \bottomrule
    \end{tabular}
    \caption{Agreement between human annotations and GPT-5.5 judgments on proposal novelty and clarity. Observed agreement measures the proportion of matching pairwise outcomes; Cohen's $\kappa$ accounts for chance agreement. CI denotes confidence interval.}
    \label{tab:idea_maker_human_agreement}
\end{table}

As shown in Table~\ref{tab:idea_maker_human_agreement}, human annotations agree with GPT-5.5 judgments in 83.94\% of novelty comparisons and 82.38\% of clarity comparisons. Both $\kappa$ estimates and their 95\% confidence intervals are above zero, indicating agreement beyond chance.

\subsection{Molecule Generator}

\paragraph{Task setting.}
The Molecule Generator benchmark instantiates the constrained inverse-design setting described in the multi-property molecular design application \cite{dey2025gellmogeneralizinglargelanguage}. Each system receives a target property profile \(Y^*\), a set of hard chemical constraints, and optional project context such as seed scaffolds, forbidden substructures, or preferred synthetic routes. The expected output is a ranked candidate set with machine-readable molecular representations and the reasoning or tool traces used to justify the ranking.

\paragraph{Dataset.}
Each benchmark task combines multiple molecular objectives, including blood-brain barrier penetration (BBBP), penalized logP (pLogP), quantitative estimate of drug-likeness (QED), mutagenicity, and human intestinal absorption (HIA). For compact reporting, we abbreviate the target combinations as BPQ for BBBP+pLogP+QED, MPQ for Mutagenicity+pLogP+QED, BHMQ for BBBP+HIA+Mutagenicity+QED, BMPQ for BBBP+Mutagenicity+pLogP+QED, and HMPQ for HIA+Mutagenicity+pLogP+QED.

\paragraph{Evaluation metrics.}
The primary metric is success rate, defined as the fraction of generated candidates that satisfy all hard constraints and meet the target property after verification. Unlike \cite{dey2025gellmogeneralizinglargelanguage}, where predicting the direction suffices, our evaluation based on target property thresholds makes the task significantly more challenging.

\paragraph{Results.}

\begin{figure}[htbp]
    \centering
\begin{tikzpicture}
    \begin{axis}[
        ybar,
        width=\textwidth,
        height=8cm,
        bar width=4pt,
        ymin=-0.015,
        ymax=0.45,
        ytick={0,0.1,0.2,0.3,0.4},
        ylabel={Success rate},
        xtick={1,2,3,4,5},
        xticklabels={BPQ, MPQ, BHMQ, BMPQ, HMPQ},
        xticklabel style={font=\small},
        legend style={at={(0.5,1.16)}, anchor=south, legend columns=3, font=\scriptsize},
        enlarge x limits=0.14,
        major grid style={dashed,gray!30},
        ymajorgrids=true,
        error bars/y dir=both,
        error bars/y explicit,
        error bars/error bar style={draw=black,line width=0.45pt},
        error bars/error mark options={rotate=90,mark size=1.2pt,line width=0.45pt}
    ]
        \addplot+[draw=SCIONBar, fill=SCIONBar!78] coordinates {(1,0.3803) +- (0,0.0023) (2,0.2720) +- (0,0.0020) (3,0.2302) +- (0,0.0301) (4,0.3342) +- (0,0.0054) (5,0.4028) +- (0,0.0267)};
        \addplot+[draw=DRClawBar, fill=DRClawBar!78] coordinates {(1,0.1763) +- (0,0.0015) (2,0.2033) +- (0,0.0012) (3,0.0367) +- (0,0.0088) (4,0.1099) +- (0,0.0159) (5,0.1528) +- (0,0.0120)};
        \addplot+[draw=AutoResearchBar, fill=AutoResearchBar!78] coordinates {(1,0.0403) +- (0,0.0006) (2,0.0460) +- (0,0.0000) (3,0.0226) +- (0,0.0065) (4,0.0227) +- (0,0.0015) (5,0.0104) +- (0,0.0000)};
        \addplot+[draw=ScienceClawBar, fill=ScienceClawBar!78] coordinates {(1,0.0977) +- (0,0.0015) (2,0.1467) +- (0,0.0015) (3,0.0282) +- (0,0.0088) (4,0.1353) +- (0,0.0223) (5,0.1493) +- (0,0.0030)};
        \addplot+[draw=AIResearcherBar, fill=AIResearcherBar!78] coordinates {(1,0.0417) +- (0,0.0006) (2,0.0480) +- (0,0.0000) (3,0.0056) +- (0,0.0024) (4,0.0166) +- (0,0.0015) (5,0.0000) +- (0,0.0000)};
        \addplot+[draw=AIScientistBar, fill=AIScientistBar!78] coordinates {(1,0.0580) +- (0,0.0010) (2,0.0597) +- (0,0.0015) (3,0.0198) +- (0,0.0024) (4,0.0419) +- (0,0.0026) (5,0.0104) +- (0,0.0000)};
        \addplot+[draw=EvoScientistBar, fill=EvoScientistBar!78] coordinates {(1,0.0720) +- (0,0.0010) (2,0.0647) +- (0,0.0012) (3,0.0056) +- (0,0.0024) (4,0.0314) +- (0,0.0026) (5,0.0226) +- (0,0.0060)};
        \addplot+[draw=InternAgentBar, fill=InternAgentBar!78] coordinates {(1,0.0593) +- (0,0.0006) (2,0.0513) +- (0,0.0006) (3,0.0198) +- (0,0.0024) (4,0.0462) +- (0,0.0015) (5,0.0208) +- (0,0.0000)};
        \addplot+[draw=ARISBar, fill=ARISBar!78] coordinates {(1,0.1317) +- (0,0.0012) (2,0.2163) +- (0,0.0012) (3,0.0494) +- (0,0.0024) (4,0.1222) +- (0,0.0040) (5,0.1024) +- (0,0.0060)};
        \legend{SCION, DR-Claw, AutoResearchClaw, ScienceClaw, AI-Researcher, AI-Scientist-v2, EvoScientist, InternAgent-1.0, ARIS + Nex-N1.1}
    \end{axis}
\end{tikzpicture}
    \caption{Success rates on multi-property molecule-generation tasks. Bars show the mean fraction of generated candidates satisfying all hard constraints and target thresholds across three independent runs; error bars show $\pm 1$ sample standard deviation. ARIS + Nex-N1.1 uses Nex-N1.1 for execution, critique, and review. Higher values indicate better performance.}
    \label{fig:moleculeoptimization}
\end{figure}

As shown in Figure~\ref{fig:moleculeoptimization}, SCION achieves the highest mean success rate on all five target combinations in the three-run comparison. Its equal-weight average success rate is 0.3239, compared with 0.1358 for DR-Claw, the strongest baseline on average. On MPQ and BHMQ, SCION reaches $0.2720 \pm 0.0020$ and $0.2302 \pm 0.0301$, respectively, compared with $0.2163 \pm 0.0012$ and $0.0494 \pm 0.0024$ for ARIS + Nex-N1.1, the strongest baseline on these tasks. On BMPQ, SCION reaches $0.3342 \pm 0.0054$, compared with $0.1353 \pm 0.0223$ for ScienceClaw. On HMPQ, SCION reaches $0.4028 \pm 0.0267$, compared with $0.1528 \pm 0.0120$ for DR-Claw.

The advantage of SCION becomes more visible when the task requires simultaneous satisfaction of multiple heterogeneous constraints. In such settings, simple generation or single-agent reasoning is insufficient: candidate molecules must be generated, checked for validity, filtered against target properties, and re-ranked under competing objectives. The higher success rates therefore support the central architectural claim of this paper: a governed agentic runtime can improve scientific generation tasks by coordinating decomposition, verification, and iterative refinement rather than relying only on one-shot model output.

\subsection{Antibody Screening}

\paragraph{Task setting.}
The antibody-screening experiment evaluates the one-round policy optimization problem introduced in the batch active search formulation \cite{Bailey_2024,gupta2025llmsbayesianoptimizationscientific,pmlr-v70-jiang17d}. At round \(t\), each system receives the accumulated screening history \(\mathcal{D}_{t-1}\), including previously tested antibody or protein candidates, binary hit labels, and numerical property feedback such as affinity, enrichment, or developability scores. The system must output a ranked list of unevaluated candidates or a batch-selection policy for the next screening round. The agents are evaluated based on synthetic feedback data obtained at an intermediate screening round t ($t>1$).

\paragraph{Dataset.}
The evaluation dataset comprises target-specific antibody candidate pools from real protein discovery campaigns, with screening labels derived from a composite property calculated from multiple quantitative attributes. We label top 10\% candidates as label 1 and the others label 0. Candidates are split evenly into training and test sets (1:1 ratio). The training set simulates the historical screening data available to the system, while the test set represents the remaining candidates to be evaluated in the current round.

\paragraph{Evaluation metrics.}
The primary metric is F1, computed by comparing the top 10\% candidates selected by the system with the true top 10\% high-performing antibodies in the test set. A higher F1 score indicates a better balance between selecting true positive candidates and avoiding low-value candidates.

\paragraph{Results.}
Figure~\ref{fig:antibodyscreening} reports the antibody-screening results across three independent evaluations of an intermediate screening round.

\begin{figure}[htbp]
    \centering
\begin{tikzpicture}
    \begin{axis}[
        ybar,
        width=\textwidth,
        height=7cm,
        bar width=10pt,
        ymin=0,
        ymax=0.42,
        ylabel={F1},
        xtick={1,2,3,4,5,6,7,8,9},
        xticklabels={SCION, DR-Claw, AutoResearchClaw, ScienceClaw, AI-Researcher, AI-Scientist-v2, EvoScientist, InternAgent-1.0, ARIS + Nex-N1.1},
        xticklabel style={rotate=35, anchor=east, font=\scriptsize},
        enlarge x limits=0.08,
        major grid style={dashed,gray!30},
        ymajorgrids=true,
        error bars/y dir=both,
        error bars/y explicit,
        error bars/error bar style={draw=black,line width=0.45pt},
        error bars/error mark options={rotate=90,mark size=1.2pt,line width=0.45pt}
    ]
        \addplot+[draw=SCIONBar, fill=SCIONBar!78, bar shift=0pt] coordinates {(1,0.3733) +- (0,0.0351)};
        \addplot+[draw=DRClawBar, fill=DRClawBar!78, bar shift=0pt] coordinates {(2,0.1033) +- (0,0.0058)};
        \addplot+[draw=AutoResearchBar, fill=AutoResearchBar!78, bar shift=0pt] coordinates {(3,0.0733) +- (0,0.0058)};
        \addplot+[draw=ScienceClawBar, fill=ScienceClawBar!78, bar shift=0pt] coordinates {(4,0.2000) +- (0,0.0200)};
        \addplot+[draw=AIResearcherBar, fill=AIResearcherBar!78, bar shift=0pt] coordinates {(5,0.2833) +- (0,0.0208)};
        \addplot+[draw=AIScientistBar, fill=AIScientistBar!78, bar shift=0pt] coordinates {(6,0.1967) +- (0,0.0058)};
        \addplot+[draw=EvoScientistBar, fill=EvoScientistBar!78, bar shift=0pt] coordinates {(7,0.1967) +- (0,0.0153)};
        \addplot+[draw=InternAgentBar, fill=InternAgentBar!78, bar shift=0pt] coordinates {(8,0.1867) +- (0,0.0058)};
        \addplot+[draw=ARISBar, fill=ARISBar!78, bar shift=0pt] coordinates {(9,0.1760) +- (0,0.0250)};
    \end{axis}
\end{tikzpicture}
    \caption{F1 results for one-round target-specific antibody screening. Bars show the mean F1 across three independent runs; error bars show $\pm 1$ sample standard deviation. ARIS + Nex-N1.1 uses the same Nex-N1.1 backbone as SCION. Higher F1 indicates better selection of high-value candidates from the held-out screening set.}
    \label{fig:antibodyscreening}
\end{figure}

SCION obtains the highest mean F1 score of $0.3733 \pm 0.0351$. The strongest baseline is AI-Researcher with $0.2833 \pm 0.0208$, giving SCION an absolute improvement of 0.0900. The matched-backbone ARIS + Nex-N1.1 configuration achieves $0.1760 \pm 0.0250$.

This result points to a stronger ability to convert historical screening feedback into an actionable next-round selection policy. It is consistent with the batch active search formulation in Section~\ref{sec:batch_active_search}: the system is not merely asked to score each candidate independently, but to use previous binary labels and numerical feedback to decide which candidates should be prioritized under a limited screening budget. The improvement in F1 therefore reflects stronger alignment between the selected candidate set and the true high-value antibody set in the held-out screening round.

\subsection{End-to-End Research Tasks}
\label{sec:rcb-ablation}

\paragraph{Task setting.}
We evaluate SCION on ResearchClawBench~\cite{xu2026researchclawbenchbenchmarkendtoendautonomous}, an end-to-end scientific research benchmark spanning ten disciplines. Each task provides related literature and raw data and requires the agent to produce scientific artifacts through literature analysis, computational execution, interpretation, and reporting. This setting closely matches SCION's role as a harness for coordinating sustained scientific workflows: progress must be preserved across stages, intermediate results must inform subsequent actions, and the outputs of different agents must be integrated into a coherent research outcome. ResearchClawBench therefore complements the application-specific evaluations by assessing how these capabilities operate together throughout an entire research process.

\paragraph{Configurations.}
We compare four cumulative configurations, each retaining the components of the preceding configuration. The research-execution backbone is held fixed across configurations. Research execution uses Kimi-K3 through Kimi Code; debugging and idea proposal use Codex with GPT-5.6 Sol.
\begin{enumerate}[leftmargin=*,itemsep=3pt,topsep=3pt]
    \item \textbf{Baseline.} The backbone performs task reasoning and tool execution through the basic agent runtime.
    \item \textbf{Hierarchical agents.} The baseline is augmented with a coordinator--worker hierarchy. The coordinator decomposes research objectives, assigns subtasks to worker agents, and integrates their outputs.
    \item \textbf{Hierarchical agents + L1/L2 memory.} L1/L2 memory is added to the hierarchical-agent configuration while retaining the coordinator--worker structure. L1 preserves recent interactions, task-local tool results, and working plans. L2 stores durable project decisions, scientific assumptions, reusable findings, and operational constraints, making accumulated knowledge available to subsequent research steps.
    \item \textbf{Full system.} This configuration uses the complete SCION auto-research system ScienceEvolve described in this paper.
\end{enumerate}

\paragraph{Results.}
Table~\ref{tab:rcb-ablation} reports the scores for all ten disciplines. Hierarchical agents obtain a macro-average of 23.57, compared with 22.39 for the baseline, an increase of 1.18 points. Adding L1/L2 memory to the hierarchical configuration raises the macro-average to 25.45, a further increase of 1.88 points and a total increase of 3.06 points over the baseline. For the full system, the mean of the highest observed score in each discipline across the evaluated evolution rounds is 29.22.

\begin{table}[htbp]
    \centering
    \small
    \caption{Incremental ablation on ResearchClawBench. Each configuration includes the components of the preceding configuration. Mean denotes the macro-average across ten disciplines. }
    \label{tab:rcb-ablation}
    \setlength{\tabcolsep}{6pt}
    \begin{tabular}{@{}lrrrr@{}}
        \toprule
        Discipline & Baseline (Kimi-K3+Kimi Code) & \makecell[r]{Hierarchical\\agents} & \makecell[r]{Hierarchical agents\\+ L1/L2 memory} & Full system \\
        \midrule
        Physics      & 26.82 & 31.10 & 32.00 & 37.10 \\
        Astronomy    & 21.12 & 24.25 & 26.82 & 29.50 \\
        Energy       & 27.74 & 27.31 & 29.49 & 32.80 \\
        Information  & 19.04 & 24.40 & 24.71 & 30.37 \\
        Chemistry    & 14.49 & 17.81 & 20.63 & 20.63 \\
        Math         & 27.67 & 21.63 & 28.24 & 28.72 \\
        Earth        & 22.96 & 21.63 & 24.24 & 35.92 \\
        Life         & 21.66 & 24.13 & 22.08 & 22.08 \\
        Neuroscience & 17.25 & 20.68 & 21.12 & 27.64 \\
        Material     & 25.17 & 22.73 & 25.19 & 27.42 \\
        \midrule
        \textbf{Mean} & \textbf{22.39} & \textbf{23.57} & \textbf{25.45} & \textbf{29.22} \\
        \bottomrule
    \end{tabular}
\end{table}

\section{Conclusion}

Every major shift in scientific productivity is accompanied by a reconstruction of the underlying infrastructure. This paper has presented SCION as an agentic scientific operating system---an organizational nexus for collaborative research---that re-architects the research process around three persistent capabilities: research intent orchestration, multi-agent execution, and plan-level knowledge reuse. With these capabilities, AI can function less as a peripheral predictive utility and more as a coordinated operational layer for end-to-end scientific work.

Equally important, SCION can shift part of the coordination burden in scientific labor. It reduces the need for human scientists to act as manual dispatchers and keeps their attention on scientific strategy, value alignment, and interpretation, while the system supports decomposition, execution, and memory formation. The application cases in materials analysis, multi-property molecular design, and target-specific antibody screening show that the significance of SCION is not merely computational scale, but the creation of a traceable and reusable mode of collaborative scientific innovation. Looking ahead, combining this operating-system model with automated experimentation and high-throughput wet labs may pave the way toward a cyber-physical infrastructure for science, which constitutes our planned future work.

\clearpage
\bibliographystyle{unsrtnat}
\bibliography{main}

\clearpage
\appendix
\section{Three-Run Evaluation Results}
\label{sec:three-run-results}
Tables~\ref{tab:reading_three_runs}--\ref{tab:antibody_three_runs} report the mean and sample standard deviation across three independent runs for each method and metric. Bold values identify the highest mean in each column. The molecule and antibody comparisons use the matched-backbone ARIS + Nex-N1.1 configuration.


\begin{table}[htbp]
    \centering
    \small
    \begin{tabular}{lcc}
        \toprule
        \textbf{Method} & \textbf{SEED} & \textbf{Accuracy (\%)} \\
        \midrule
        \textbf{SCION} & $\mathbf{44.02 \pm 0.54}$ & $\mathbf{35.90 \pm 0.29}$ \\
        DR-Claw & $34.32 \pm 0.32$ & $26.22 \pm 0.29$ \\
        AutoResearchClaw & $26.51 \pm 1.03$ & $22.05 \pm 0.40$ \\
        ScienceClaw & $21.94 \pm 0.77$ & $16.67 \pm 0.29$ \\
        AI-Researcher & $38.66 \pm 0.26$ & $30.13 \pm 0.22$ \\
        AI-Scientist-v2 & $38.28 \pm 0.16$ & $29.74 \pm 0.11$ \\
        EvoScientist & $42.40 \pm 0.62$ & $34.04 \pm 0.58$ \\
        InternAgent-1.0 & $37.29 \pm 1.56$ & $29.36 \pm 0.48$ \\
        \bottomrule
    \end{tabular}
    \caption{Scientific Reading results on the 520 CMPhysBench questions. Values are mean $\pm$ sample standard deviation over three independent runs. All methods use Kimi-K2.5.}
    \label{tab:reading_three_runs}
\end{table}

\begin{table}[htbp]
    \centering
    \small
    \setlength{\tabcolsep}{3pt}
    \resizebox{\textwidth}{!}{%
    \begin{tabular}{lccccc}
        \toprule
        \textbf{Method} & \textbf{BPQ} & \textbf{MPQ} & \textbf{BHMQ} & \textbf{BMPQ} & \textbf{HMPQ} \\
        \midrule
        \textbf{SCION} & $\mathbf{0.3803 \pm 0.0023}$ & $\mathbf{0.2720 \pm 0.0020}$ & $\mathbf{0.2302 \pm 0.0301}$ & $\mathbf{0.3342 \pm 0.0054}$ & $\mathbf{0.4028 \pm 0.0267}$ \\
        DR-Claw & $0.1763 \pm 0.0015$ & $0.2033 \pm 0.0012$ & $0.0367 \pm 0.0088$ & $0.1099 \pm 0.0159$ & $0.1528 \pm 0.0120$ \\
        AutoResearchClaw & $0.0403 \pm 0.0006$ & $0.0460 \pm 0.0000$ & $0.0226 \pm 0.0065$ & $0.0227 \pm 0.0015$ & $0.0104 \pm 0.0000$ \\
        ScienceClaw & $0.0977 \pm 0.0015$ & $0.1467 \pm 0.0015$ & $0.0282 \pm 0.0088$ & $0.1353 \pm 0.0223$ & $0.1493 \pm 0.0030$ \\
        AI-Researcher & $0.0417 \pm 0.0006$ & $0.0480 \pm 0.0000$ & $0.0056 \pm 0.0024$ & $0.0166 \pm 0.0015$ & $0.0000 \pm 0.0000$ \\
        AI-Scientist-v2 & $0.0580 \pm 0.0010$ & $0.0597 \pm 0.0015$ & $0.0198 \pm 0.0024$ & $0.0419 \pm 0.0026$ & $0.0104 \pm 0.0000$ \\
        EvoScientist & $0.0720 \pm 0.0010$ & $0.0647 \pm 0.0012$ & $0.0056 \pm 0.0024$ & $0.0314 \pm 0.0026$ & $0.0226 \pm 0.0060$ \\
        InternAgent-1.0 & $0.0593 \pm 0.0006$ & $0.0513 \pm 0.0006$ & $0.0198 \pm 0.0024$ & $0.0462 \pm 0.0015$ & $0.0208 \pm 0.0000$ \\
        ARIS + Nex-N1.1 & $0.1317 \pm 0.0012$ & $0.2163 \pm 0.0012$ & $0.0494 \pm 0.0024$ & $0.1222 \pm 0.0040$ & $0.1024 \pm 0.0060$ \\
        \bottomrule
    \end{tabular}
    }
    \caption{Multi-property molecule-generation success rates, reported as mean $\pm$ sample standard deviation over three independent runs. ARIS + Nex-N1.1 uses Nex-N1.1 for execution, critique, and review.}
    \label{tab:molecule_three_runs}
\end{table}

\begin{table}[htbp]
    \centering
    \small
    \begin{tabular}{lc}
        \toprule
        \textbf{Method} & \textbf{F1} \\
        \midrule
        \textbf{SCION} & $\mathbf{0.3733 \pm 0.0351}$ \\
        DR-Claw & $0.1033 \pm 0.0058$ \\
        AutoResearchClaw & $0.0733 \pm 0.0058$ \\
        ScienceClaw & $0.2000 \pm 0.0200$ \\
        AI-Researcher & $0.2833 \pm 0.0208$ \\
        AI-Scientist-v2 & $0.1967 \pm 0.0058$ \\
        EvoScientist & $0.1967 \pm 0.0153$ \\
        InternAgent-1.0 & $0.1867 \pm 0.0058$ \\
        ARIS + Nex-N1.1 & $0.1760 \pm 0.0250$ \\
        \bottomrule
    \end{tabular}
    \caption{Antibody-screening F1, reported as mean $\pm$ sample standard deviation over three independent runs.}
    \label{tab:antibody_three_runs}
\end{table}

\end{document}